\documentclass{article}

\usepackage{iclr2027_conference,times}

\usepackage{hyperref}
\usepackage{url}
\usepackage{subcaption}
\usepackage{booktabs}
\usepackage{multirow}
\usepackage[utf8]{inputenc}
\usepackage[T1]{fontenc}
\usepackage{amsfonts}
\usepackage{nicefrac}
\usepackage{microtype}
\usepackage[table,xcdraw]{xcolor}
\usepackage{graphics}
\usepackage{graphicx}
\usepackage{multicol}
\usepackage{enumitem}
\usepackage{wrapfig}
\usepackage{titlesec}
\usepackage{listings}
\usepackage{fvextra}
\usepackage{makecell}
\usepackage{amssymb}
\usepackage{amsmath}
\usepackage{titletoc}
\usepackage{fancyhdr}

\title{Fidel-TS: A High-Fidelity Multimodal Benchmark for Time Series Forecasting}

\author{
  Zhijian Xu\textsuperscript{1}\thanks{Equal contribution.} \quad
  Wanxu Cai\textsuperscript{1}\footnotemark[1] \quad
  Xilin Dai\textsuperscript{1} \quad
  Zhaorong Deng\textsuperscript{2} \quad
  Qiang Xu\textsuperscript{1}\thanks{Corresponding author.} \\
  \\ 
  \textsuperscript{1}Department of Computer Science and Engineering, The Chinese University of Hong Kong \\
  \textsuperscript{2}School of Data Science, The Chinese University of Hong Kong, Shenzhen \\
  \texttt{zjxu21@cse.cuhk.edu.hk, wxcai26@cse.cuhk.edu.hk} \\
  \texttt{xldai26@cse.cuhk.edu.hk, 122090081@link.cuhk.edu.cn} \\
  \texttt{qxu@cse.cuhk.edu.hk}
}

\iclrfinalcopy

\begin{document}

\maketitle

\pagestyle{fancy}
\fancyhf{}
\fancyhead[C]{\small Preprint}


\begin{abstract}
The evaluation of time series forecasting models is hindered by a lack of high-quality benchmarks, leading to overestimated assessments of progress. Existing datasets suffer from issues ranging from small-scale, low-frequency, pre-training data contamination in unimodal designs to temporal and description leakage prevalent in early multimodal designs. To address this, we formalize the core principles of high-fidelity benchmarking, focusing on data sourcing integrity, leak-free design, and structural clarity. We introduce Fidel-TS, a new large-scale benchmark built from these principles. Our experiments reveal the limitations of prior benchmarks and the potential discrepancies in model evaluation, providing new insights into multiple existing unimodal and multimodal forecasting models and LLMs across various evaluation tasks. The code is available at \href{https://github.com/VEWOXIC/Universal-Cross-Modal-Time-Series-Forecasting-Pipeline}{https://github.com/VEWOXIC/Universal-Cross-Modal-Time-Series-Forecasting-Pipeline} and data is available at \href{https://huggingface.co/collections/fidel-ts/fidel-ts}{https://huggingface.co/collections/fidel-ts/fidel-ts}.
\end{abstract}


\section{Introduction}
\label{sec:introduction}

The field of time series forecasting has seen relentless progress, with a proliferation of methods from classic statistical models \citep{box2015time} and machine learning models \citep{chen2016xgboost} to sophisticated deep learning architectures \citep{hochreiter1997long, vaswani2017attention} and foundation models \citep{garza2023timegpt, rasul2023lag}. As models demonstrate increasingly strong capabilities, the field of time-series forecasting calls for evaluations that are broader, fairer, and more aligned with real-world scenarios. However, we argue that the benchmarks currently in use have not kept pace with the rapid advancements in model development.

This foundation faces both historical limitations and new methodological challenges. Previous unimodal benchmarks \citep{zhou2021informer, wu2021autoformer, wu2023timesnet} are often small in scale, outdated, and feature a vague composition of variables, making them less suitable for robustly evaluating modern models. More critically, the widespread availability of these datasets has created a significant, often unacknowledged, risk of pre-training contamination for large language models (LLMs) \citep{magar2022data, jiang2024does}, which may lead to an overestimation of LLMs' capabilities in time series forecasting. As the community has pivoted to multimodal forecasting \citep{williams2024context, wang2024news, xu2024intervention}, first-generation benchmarks \citep{liu2024time} have inherited this contamination risk while encountering additional methodological challenges. The common practice of retrospectively retrieving textual data introduces severe data leakage, including temporal leakage (accessing information published after the event to be predicted) and description leakage (retrieving text that explicitly states the ground-truth value). Furthermore, relying on low-frequency data (e.g., monthly) limits the data volume, constraining the effective training of domain-specific numerical models and prevents them from providing complex testing scenarios.

To rectify these shortcomings and establish a more credible evaluation standard, we reconsider the essential characteristics of data encountered in real-world forecasting scenarios. Based on this review, we formalize core principles of \textbf{high-fidelity benchmarking}: (1) \textbf{Data Sourcing Integrity}, which mandates the use of real-time, authentication-protected API data streams to not only ensure recency and data sufficiency but also mitigate pre-training contamination; (2) \textbf{Leak-free Design}, which incorporates only verifiably scheduled exogenous textual information, including weather forecasts and control events, to prevent temporal and description leakage; (3) \textbf{Structural Clarity}, which implements a clear demarcation between forecasting "Subjects" and data "Channels" to enable rigorously evaluating model generalization. Guided by these principles, we introduce \textbf{Fidel-TS}, a new benchmark designed from the ground up to embody these principles. Each dataset contains large scale of fresh, high-frequency and leak-free data points from real-time APIs, features textual information covering every timestamp, and is built with a highly extensible architecture.

To demonstrate the value of our high-fidelity benchmark, we developed a comprehensive framework compatible with models of diverse modalities and architectures. This assessment reveals a performance gap compared to earlier reports, exposing the evaluation biases inherent in prior benchmarks. From this evaluation, we obtain three key findings. First, in unimodal forecasting, no model consistently dominates, and Time Series Foundation Models fail to exhibit the claimed robust zero-shot forecasting ability. Second, multimodal gains depend on model architecture and the relevance of textual inputs. Third, although LLM performance correlates with general reasoning ability, even domain-finetuned variants remain unstable; under realistic, context-rich, and leakage-free long-term settings, LLMs employing direct prompting suffer from severe precision and reliability issues and generally underperform dedicated forecasting models.

In summary, our contributions are:
\begin{itemize}
    \item A formalization of the principles for high-fidelity time series benchmarking.
    \item The release of Fidel-TS, a new large-scale benchmark built on these principles.
    \item A thorough experimental analysis using our proposed framework, offering new insights into the capabilities of modern forecasting models.
\end{itemize}

\section{Related Works}
\label{sec:related works}

\subsection{Unimodal Benchmarks for Time Series Forecasting} The evaluation of time series forecasting models has historically relied on a narrow set of previous unimodal benchmarks, typically including ETT(ETTh1, ETTh2, ETTm1, ETTm2), ECL \citep{zhou2021informer}; Electricity, Weather, Traffic, ILI \citep{wu2021autoformer}; M4 \citep{wu2023timesnet}, etc. The utility of these benchmarks is constrained by three main factors:

First, they are \textbf{small in scale}, with most datasets containing only on the order of tens of thousands of data points. This limited data volume is often insufficient for robust model evaluation, meaning that observed performance gains may be statistically insignificant or biased.

Second, they are \textbf{outdated} and have been widely accessible online for years. For instance, the ETT dataset concluded in 2018 and the Electricity dataset concluded in 2019. This long-term availability poses a significant risk of pre-training contamination, particularly for LLMs, casting doubt on the fairness of evaluations on modern LLM-based models.

Third, their \textbf{variable composition lacks explicit structural demarcation}. There is no clear demarcation between datasets containing multiple variables from a single system and those featuring a single variable type across multiple independent systems. For example, the Weather dataset comprises various meteorological variables for atmosphere, whereas the Electricity/Traffic datasets treat sensors from different locations as distinct variables, despite them all measuring the same physical quantity. 

\subsection{Multimodal Benchmarks for Time Series Forecasting} While the shift to multimodality was conceptually correct, transitioning from unimodal to multimodal benchmarks is a non-trivial challenge. Previous unimodal datasets, due to their age and lack of rich contextual metadata, are difficult to extend. The initial multimodal time series forecasting benchmark, TimeMMD \citep{liu2024time}, addressed this problem but highlighted the challenges of maintaining real world fidelity, demonstrated several limitations:

First, the majority of its datasets are plagued by \textbf{low sampling frequencies}. Six datasets are sampled monthly, including Agriculture, Climate, Economy, Security, Social Good, and Traffic, while Energy and Health are sampled weekly. This results in a severe scarcity of data points, with many series containing only a few hundred observations despite spanning decades. These simple samples cannot support the evaluation of complex scenarios or the training of numerical domain-specific models.

Second, the reliance on an agent pipeline to retrieve web-based text introduces multiple potential forms of data leakage. This method risks \textbf{pre-training contamination}, as static web content is likely already part of the training corpora for LLMs. It also leads to \textbf{temporal leakage}, where retrieved documents like future financial reports contain information unavailable at prediction time, and \textbf{description leakage}, where text explicitly describes future trends, effectively revealing the ground truth. Furthermore, this approach fails to ensure strict temporal alignment between the textual data and its corresponding time series timestamp, increasing the risk of temporal leakage.

Third, it perpetuates the \textbf{unclear variable structure} of unimodal benchmarks. It also fails to distinguish between datasets comprising multiple, distinct variables from a single system versus those with a single variable type from multiple systems. Besides, in its multivariate datasets, it arbitrarily designates one channel as the "target," a design choice that may oversimplify the problem and bypass the potential interdependencies among variables.

Taken together, these characteristics regarding data scale, potential leakage, and structural ambiguity can limit the real-world fidelity of the benchmarks, thereby affecting the evaluation of the models.

\section{Fidel-TS: A Benchmark Built on High-Fidelity Principles}
\label{sec:datasets}

\subsection{Core Principles of High-Fidelity Design}

\paragraph{Authentication-Protected API Data Streams} At the core of our benchmark is the exclusive utilization of continuously updated data streams from authentication-protected APIs, representing a significant departure from the static, web-scraped files common in existing benchmarks. This architectural design fundamentally enhances the realism of our evaluation. It supplies a massive volume of fresh, high-frequency data, ensuring the benchmark remains capable of assessing fine-grained, sub-hourly tasks while maintaining the precise cross-modal temporal alignment critical for accurate forecasting. Furthermore, this API-driven approach inherently constructs a defense against data contamination. Sourcing data from restricted APIs rather than widely-indexed public repositories minimizes its presence in common training corpora.

\paragraph{Scheduled Data as Textual Modality} We deliberately eschew open-ended textual sources from previous works \textbf{(including news, public events, and political or economic reports)} \citep{wang2024news, williams2024context, liu2024time} in favor of scheduled data (including \textbf{weather forecasts} and dataset-specific \textbf{control events}). First, in real-world online forecasting, predictions must rely strictly on information available at the forecasting timestamp. Scheduled data uniquely satisfies this requirement: weather forecasts are released on fixed schedules, and planned control events are known before execution, making their availability explicit and verifiable. In contrast, open-ended sources such as news are inherently unpredictable, as their publication timing cannot be assumed ex ante, introducing temporal leakage risks; they may also explicitly state future outcomes (e.g., ``Energy supply will be increased''), causing description leakage. Second, news typically exhibits low-frequency or irregular updates that fail to align with high-frequency time series. Scheduled data instead provides structured, timestamped signals that naturally align with granular sub-hourly sequences. Furthermore, scheduled data is inherently causally grounded. As demonstrated in domains like photovoltaics, electricity, and transportation \citep{antonanzas2016review, hong2016probabilistic, polson2017deep}, these variables are widely acknowledged as vital indicators that actively shape future system dynamics, rather than merely serving as historical covariates.


\paragraph{Structural Distinction Between Subjects and Channels}
To accurately mirror real-world data collection, we formalize a precise structural distinction between subjects and channels. Across our diverse sources, datasets are systematically organized where data originates from multiple distinct subjects (e.g., specific regions, devices, or sensors), with each subject recording an identical set of channels (i.e., the observed variables). For instance, in an electricity dataset, individual households represent the subjects, whereas specific measurements like 'usage (kWh)' and 'voltage (V)' constitute the channels. This explicit architectural separation is critical for rigorously evaluating model generalization, as it faithfully reflects practical deployment scenarios, such as applying a trained model to a newly installed sensor within an existing system.


\paragraph{Extensibility and Reproducibility} Regarding continuous updates, our data construction pipeline enables researchers to dynamically fetch the latest data via real-time APIs, while archiving all historical records to guarantee complete experimental reproducibility. Regarding structural expansion, this flexible workflow empowers researchers to easily incorporate custom time series and heterogeneous data modalities, ensuring precise temporal alignment.


\begin{figure*}
    \centering
    \includegraphics[width=\linewidth]{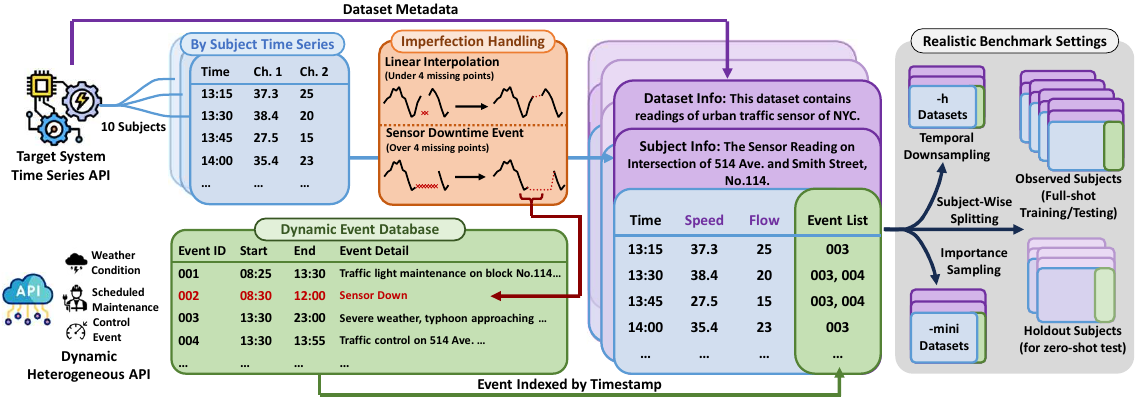}
    \caption{\textbf{The Construction Pipeline of Fidel-TS}. The workflow integrates raw data from the Target System Time Series API and the Dynamic Heterogeneous API. The Imperfection Handling step interpolates short data gaps while converting long downtime into time-aligned 'Sensor Downtime'. These downtime events and heterogeneous textual data are stored in the Dynamic Event Database together, strictly following the start and end times. All these events and time series, combined with static dataset metadata, forms a unified dataset that is systematically organized into Realistic Benchmark Settings: downsampled (\texttt{-h}), sampled via importance (\texttt{-mini}), observed (\texttt{-obs}, for in-domain evaluation), and hidden (\texttt{-hid}, for generalization evaluation).}
    \vspace{-0.2cm}
    \label{fig:pipeline}
\end{figure*}

\subsection{Data Curation Pipeline}

Translating raw, real-time data streams into a structured benchmark requires a pipeline that preserves the nuances of real-world data. The following steps detail our process, with a full summary of the complete pipeline in Figure \ref{fig:pipeline} and the resulting datasets in Table \ref{tab:all datasets}. More detailed background information of these datasets is provided in Appendix \ref{sec:details of datasets}.

\paragraph{Handling Real-World Data Imperfections.} Embodying our principle of high fidelity, a core feature of the benchmark data is that it directly reflects the challenges of real-world instrumentation. Sourced from physical sensors rather than statistically processed tables, the raw time series data naturally contains imperfections like environmental noise and missing values. This inherent instability is a feature, not a flaw, representing the unfiltered reality of live data streams. Our first step involves filtering obvious outliers and aligning timestamps to a consistent sampling rate. To handle missing values realistically, we distinguish between short and long gaps. Short-duration gaps are filled using linear interpolation, reflecting minor data transmission issues. However, longer gaps are treated as significant real-world events; we mark the numerical values as zero and generate a corresponding textual description, such as 'sensor downtime', which is aligned with the relevant timestamps. This transforms a data quality issue into a valuable contextual feature for the models to learn from. Note that long-term gaps account for only a small proportion of our data and thus introduce negligible bias to the evaluation. Detailed statistics are provided in Appendix \ref{sec: missing handling}.

\paragraph{Enriching Data with Static and Dynamic Textual Context.} Real-world forecasting requires both foundational knowledge and real-time updates. To provide this, we integrate two distinct types of textual information. First, we establish the static context by sourcing background knowledge from the official API metadata: a general overview for the entire dataset, specific descriptions for each Subject, and detailed explanations for each Channel. Second, we incorporate dynamic, time-aligned information (including weather and control events) that reflects the evolving state of the system and its environment. This dynamic information, along with detected sensor downtime events in Imperfection Handling step, is consolidated into the Dynamic Event Database, which features storage indexed by start and end times, ensuring that the textual context is strictly aligned with the timestamps of the numerical time series. See Appendix \ref{sec:visual} for examples and visual analysis of our preprocessed time series and textual data.


\paragraph{Designing Evaluation Scenarios.} To test model performance under diverse real-world conditions, we create several dataset variations. First, to ensure a fair comparison across datasets with different native sampling frequencies, we provide an option to downsample high-frequency series to a uniform hourly level (denoted by the \texttt{-h} suffix). Second, to address the unique computational challenges of evaluating LLMs, we curate smaller subsets (denoted by the \texttt{-mini} suffix) constructed via importance sampling. By selecting samples from the full test set where unimodal models outperform multimodal ones and vice versa, we ensure both efficiency and fairness. We provide more details of the algorithm in Appendix \ref{sec:importance sampling} and prove that this method does not introduce bias to the evaluation.

\paragraph{Subsetting for Generalization and Transferability.} A crucial aspect of fidelity is a model's ability to generalize to new scenarios it has not been explicitly trained on. Our Subject and Channel definitions naturally support this evaluation. For datasets with a large number of Subjects, we create distinct partitions to simulate a subject-level generalization evaluation. Subsets intended for standard training and evaluation are marked with the \texttt{-obs} (observed) suffix, while those held back for testing generalization on unseen subjects are marked with the \texttt{-hid} (hidden) suffix, providing a robust test of a model's transfer learning capabilities. More details of the division are provided in Appendix \ref{sec:subsetting}.



\begin{table*}[]
  \centering
  \caption{Overview of Dataset Statistics in Fidel-TS}
  \label{tab:all datasets}
  \resizebox{\textwidth}{!}{%
\begin{tabular}{cccccccc}
\toprule
Dataset                                                                        & Time Span                                                                                             & \begin{tabular}[c]{@{}c@{}}Sampling \\ Frequency \end{tabular} & \begin{tabular}[c]{@{}c@{}} No.\\ Subject \end{tabular} & \begin{tabular}[c]{@{}c@{}}No. Channel  \\(per Subject)\end{tabular} & \begin{tabular}[c]{@{}c@{}}Textual Data\\ Type\end{tabular}        & \begin{tabular}[c]{@{}c@{}}Textual Data\\ Frequency\end{tabular}  & \begin{tabular}[c]{@{}c@{}}Total\\Data Points \end{tabular}               \\ \midrule
\textbf{\begin{tabular}[c]{@{}c@{}}Canada Calgary \\ Photovoltaics (CCP)\end{tabular}} & \begin{tabular}[c]{@{}c@{}} 2018-01-01 \\$\sim$ 2025-02-28  \end{tabular} & 1h                                                           & 9                                                       & 1                                                                     & Weather                                                          & 6h, 1day     &  564,881                                                            \\ \midrule
\textbf{\begin{tabular}[c]{@{}c@{}}Germany Renewable\\ Energy Grid (GREG)\end{tabular}} & \begin{tabular}[c]{@{}c@{}} 2015-01-01 \\$\sim$ 2025-02-25 \end{tabular}& 15min                                                        & 8                                                       & 1                                                                      & Weather                                                            & 6h, 1day        &   2,848,472                                                           \\\midrule
\textbf{\begin{tabular}[c]{@{}c@{}}Jena Atmospheric\\ Physics (JAP)\end{tabular}}    & \begin{tabular}[c]{@{}c@{}} 2014-01-01 \\$\sim$ 2025-06-30    \end{tabular}        & 10min                                                        & 1                                                       & 21                                                                      & Weather                                                           & 6h, 1day  & 604,311                                                               \\\midrule
\textbf{\begin{tabular}[c]{@{}c@{}}NYC Traffic  \\Speed (NYTS)\end{tabular}}                                                     & \begin{tabular}[c]{@{}c@{}} 2018-01-01 \\$\sim$ 2025-03-01  \end{tabular}  & 5min                                                         & 97                                                       & 1                                                                     & Weather                                                           & 6h, 1day   & 67,517,110                                                              \\\midrule
\textbf{\begin{tabular}[c]{@{}c@{}}California \\ISO (CAISO)\end{tabular}}                                                        & \begin{tabular}[c]{@{}c@{}} 2018-04-10 \\$\sim$ 2025-02-28   \end{tabular}      & 5min                                                         & 1                                                      & 20                                                                     & Weather                                                           & 6h, 1day  & 724,896                                                               \\\midrule
\textbf{\begin{tabular}[c]{@{}c@{}}UCSD BEAR\\ Room (BEAR)\end{tabular}}                                                             & \begin{tabular}[c]{@{}c@{}} 2020-06-01 \\$\sim$ 2020-10-20    \end{tabular}       & 5min                                                         & 80                                                      & 3                                                                      & \begin{tabular}[c]{@{}c@{}} Control Events\end{tabular} & \begin{tabular}[c]{@{}c@{}}5min \end{tabular} & 3,156,240
\\
\bottomrule
\end{tabular} 
  }
  \vspace{-0.2cm}
\end{table*}

\section{Experiments}
\label{sec:experiments}

\subsection{Evaluation Framework}

Existing toolkits (including LSTF-Linear \citep{zeng2023transformers}, TSLib \citep{wang2024tssurvey}, MM-TSFLib \citep{liu2024time}, TSFM-Bench \citep{li2025tsfm}) often tailored to specific model categories and lack the flexibility to accommodate this wide spectrum of models under a single, unified configuration. To bridge the gap, we developed our \textbf{Unified Cross-modal Evaluation Framework} designed to connect our high-fidelity data with the diverse model ecosystem. The comparison in Table \ref{tab:framework} illustrates the comprehensiveness of our benchmark across training paradigms, model types, modality support, and evaluation strategies. Our framework provides an integrated testing environment for diverse models and facilitates unified experimental configuration. More details of the implementation, including the training and inference framework, the splitting and subsetting methods, and the timestamp alignment strategies, are provided in Appendix \ref{sec:framework details}.

\subsection{Experimental Setup}

\begin{wraptable}{r}{0.5\textwidth}
\centering
\vspace{-0.4cm}
\setlength{\tabcolsep}{3.5pt}
\caption{Comparison of forecasting frameworks}
\label{tab:framework}
\resizebox{\linewidth}{!}{%
\begin{tabular}{llccccc}
\toprule
\multicolumn{2}{c}{\multirow{3}{*}{\textbf{Attributes}}} & \multicolumn{5}{c}{\textbf{Benchmark Frameworks}} \\
\cmidrule(l){3-7} 
\multicolumn{2}{c}{} & \makecell{LSTF-\\Linear} & TSLib & \makecell{TSFM-\\Bench} & \makecell{MM-\\TSFLib} & \textbf{Ours} \\ 
\midrule
\multirow{2}{*}{Model} & Dom.-Spec. & $\checkmark$ & $\checkmark$ & $\checkmark$ & $\checkmark$ & $\checkmark$ \\
 & Foundation & & & $\checkmark$ & & $\checkmark$ \\ 
\midrule
\multirow{3}{*}{Structure} & Numerical & $\checkmark$ & $\checkmark$ & $\checkmark$ & $\checkmark$ & $\checkmark$ \\
 & \makecell[l]{LLM-based} & & & $\checkmark$ & & $\checkmark$ \\
 & LLM APIs & & & & $\checkmark$ & $\checkmark$ \\ 
\midrule
\multirow{2}{*}{Modality} & Unimodal & $\checkmark$ & $\checkmark$ & $\checkmark$ & $\checkmark$ & $\checkmark$ \\
 & Multimodal & & & & $\checkmark$ & $\checkmark$ \\ 
\midrule
\multirow{2}{*}{Setting} & Full-shot & $\checkmark$ & $\checkmark$ & $\checkmark$ & $\checkmark$ & $\checkmark$ \\
 & Zero-shot & & & $\checkmark$ & & $\checkmark$ \\ 
\bottomrule
\end{tabular}
}
\vspace{-0.2cm}
\end{wraptable}


\paragraph{Model Selection} Using our newly-proposed framework, we conduct a comprehensive evaluation across representative models, which we broadly categorize into:

(1) \textbf{Unimodal Forecasting Models}, including MLP-based DLinear \citep{zeng2023transformers}, FITS \citep{xu2023fits}, transformer-based PatchTST \citep{nie2023patch}, iTransformer \citep{liu2023itransformer}, LLM-reprogramming-based GPT4TS \citep{zhou2023one}, and advanced foundation models Chronos \citep{ansari2024chronos}, TimeMoE \citep{shitime} and Sundial \citep{liu2025sundial}, which are pre-trained on vast numerical time series corpora.

(2) \textbf{Multimodal Forecasting Models}, including LLM-reprogramming-based GPT4MTS \citep{jia2024gpt4mts} and transformer-based FIATS \citep{xu2024intervention}. They have different textual input requirements due to their distinct modeling approaches, representing the two main categories of multimodal forecasting models.

(3) \textbf{LLMs}, including general-purpose LLMs Qwen-2.5-14B-Instruct \citep{Yang2024Qwen25TR}, Qwen-2.5-14B-Instruct-1M \citep{yang2025qwen2.51m}, Qwen-3-14B \citep{yang2025qwen3}, Deepseek-R1 \citep{guo2025deepseek}, and expert LLM Chattime \citep{wang2025chattime} which fine-tuned based on Llama-2-7b \citep{touvron2023llama}.

See Appendix \ref{sec:model} for more details of models and Appendix \ref{sec:computation} for computational resources statistics.

\paragraph{Evaluation Metrics} We adopt the \textbf{Mean Squared Error (MSE)} and \textbf{Mean Absolute Error (MAE)} as the main evaluation metric.
Furthermore, we add a pass rate to evaluate the LLMs, defined identically to the \textbf{pass@3} metric used for tasks like code generation or math problem-solving. This addresses cases where the LLM occasionally disregard prompt constraints, causing format errors (e.g., incorrect prediction length or variable numbers) and task failure. See Appendix~\ref{sec:metrics} for formulas.

\paragraph{Realistic Settings} To simulate diverse real-world time scales, our design aligns forecasting horizons with intuitive and operational periods. For instance, for datasets with an hourly frequency, we employ historical horizon of 360 (indicates half a month) and prediction lengths of \{24, 168, 336, 720\} which correspond to \{1 day, 1 week, 2 weeks, and 1 month\}. Our framework not only supports classic ratio-based splitting (e.g., 7:1:2) \citep{zhou2021informer, wu2021autoformer}, but also enables splitting by timestamps (e.g. Jan. 1st, 2021, Jan. 1st, 2022). This approach makes the evaluation setup more transparent and controllable. See Appendix \ref{sec:settings} for more details of settings.

To guide our experiments, we formulate four \textbf{Research Questions (RQs)} and evaluate different types of models on various forecasting tasks.

\subsection{RQ1: How do unimodal forecasting models perform on the new benchmark?}

\begin{table}[ht]
\centering
\LARGE

\vspace{-0.2cm}
\caption{Unimodal forecasting performance on the full test set of each dataset. All metrics are MSE. Domain-specific models are trained on training set of each dataset, while foundation models are evaluated in zero-shot forecasting without fine-tuning on training sets. Best results are shown in \textbf{bold}.}
\label{tab:baseline unimodal}
\setlength{\tabcolsep}{1.8pt}
\scriptsize
\resizebox{\linewidth}{!}{
\begin{tabular}{l|cccc|cccc|cccc|cccc|cccc|cccc}
\toprule
& \multicolumn{4}{c|}{CCP-h}
& \multicolumn{4}{c|}{GREG-h}
& \multicolumn{4}{c|}{JAP-h}
& \multicolumn{4}{c|}{CAISO-h}
& \multicolumn{4}{c|}{NYTS-h-obs}
& \multicolumn{4}{c}{BEAR-obs} \\
\textbf{Model}
& 24 & 168 & 336 & 720
& 24 & 168 & 336 & 720
& 24 & 168 & 336 & 720
& 24 & 168 & 336 & 720
& 24 & 168 & 336 & 720
& 12 & 144 & 288 & 576 \\
\midrule

\multicolumn{25}{l}{\textit{Domain-Specific Models}} \\
\midrule

Patch.
& 0.159 & 0.205 & 0.224 & 0.258
& 0.094 & \textbf{0.130} & 0.139 & 0.163
& 0.362 & 0.541 & 0.591 & 0.663
& 0.183 & \textbf{0.360} & \textbf{0.429} & \textbf{0.536}
& 0.601 & \textbf{1.026} & \textbf{1.176} & \textbf{1.275}
& 0.121 & 0.483 & 0.680 & 0.874 \\

iTrans.
& \textbf{0.147} & \textbf{0.186} & \textbf{0.197} & \textbf{0.218}
& 0.090 & 0.132 & \textbf{0.138} & \textbf{0.155}
& \textbf{0.356} & 0.546 & 0.593 & 0.663
& 0.186 & 0.386 & 0.452 & 0.555
& \textbf{0.581} & 1.041 & 1.195 & 1.295
& \textbf{0.105} & \textbf{0.416} & \textbf{0.619} & \textbf{0.802} \\

DLinear
& 0.158 & 0.205 & 0.221 & 0.253
& 0.100 & 0.151 & 0.172 & 0.199
& 0.376 & 0.544 & \textbf{0.586} & \textbf{0.647}
& 0.190 & 0.392 & 0.486 & 0.655
& 0.638 & 1.098 & 1.262 & 1.378
& 0.128 & 0.506 & 0.650 & 0.837 \\

FITS
& 0.161 & 0.205 & 0.223 & 0.254
& 0.099 & 0.131 & 0.141 & 0.167
& 0.382 & 0.553 & 0.599 & 0.674
& 0.193 & 0.368 & 0.437 & 0.550
& 0.644 & 1.125 & 1.306 & 1.457
& 0.169 & 0.582 & 0.767 & 1.039 \\

GPT4TS
& 0.149 & 0.193 & 0.206 & 0.236
& 0.088 & 0.131 & 0.139 & 0.157
& 0.358 & \textbf{0.540} & 0.591 & 0.668
& 0.183 & 0.366 & 0.444 & 0.542
& 0.582 & 1.065 & 1.194 & 1.329
& 0.113 & 0.474 & 0.652 & 0.937 \\

\midrule
\multicolumn{25}{l}{\textit{Foundation Models}} \\
\midrule

Chronos
& 0.152 & 0.205 & 0.229 & 0.288
& \textbf{0.084} & 0.137 & 0.152 & 0.188
& 0.360 & 0.670 & 0.723 & 0.820
& 0.183 & 0.389 & 0.473 & 0.608
& 0.613 & 1.177 & 1.512 & 1.799
& 0.151 & 1.056 & 1.312 & 1.672 \\

T.-MoE
& 0.168 & 0.218 & 0.246 & 0.362
& 0.095 & 0.144 & 0.164 & 0.260
& 0.359 & 0.565 & 0.630 & 0.735
& \textbf{0.177} & 0.377 & 0.464 & 0.610
& 0.680 & 1.210 & 1.421 & 1.592
& 0.147 & 0.765 & 1.012 & 1.299 \\

Sundial
& 0.167 & 0.227 & 0.246 & 0.275
& 0.090 & 0.142 & 0.151 & 0.172
& 0.361 & 0.592 & 0.634 & 0.706
& 0.181 & 0.382 & 0.460 & 0.577
& 0.649 & 1.275 & 1.497 & 1.646
& 0.141 & 0.787 & 1.043 & 1.353 \\

\bottomrule
\end{tabular}
}
\end{table}

\paragraph{Inconsistent superiority across datasets} To explore the ranking of the capabilities of the selected unimodal forecasting models on our benchmark, we evaluated all of them and summarized the results in Table \ref{tab:baseline unimodal}. Although many domain-specific models previously reported state-of-the-art performance on classical unimodal benchmarks, their effectiveness on our benchmark is highly dataset-dependent, with different models achieving the best results on different datasets. This suggests that their reported superiority is not universally robust and may partially reflect benchmark-specific biases. Appendix \ref{sec:addtional without downsample} reports results at the original sampling frequencies, which lead to the same conclusions, while Appendix \ref{sec:all mae} presents results using Mean Absolute Error (MAE), further confirming that our findings are not sensitive to the choice of evaluation metric.

\paragraph{The failure of TSFM zero-shot forecasting} Despite claims that large-scale foundation models can generalize across diverse time series in a zero-shot manner, their performance in Table \ref{tab:baseline unimodal} is only comparable to specialized models on short forecast horizons and deteriorates substantially as the horizon extends. This indicates that the superior zero-shot capabilities reported on prior benchmarks fail to materialize our evaluation conditions.

\subsection{RQ2: How do models perform under our new generalization test?}

\begin{table}[ht]
\centering
\caption{Prior and new generalization test of domain-specific unimodal forecasting models. All metrics are MSE. We report the difference in MSE between generalization tests and in-domain evaluations in Table \ref{tab:baseline unimodal}; a smaller MSE gap indicates a closer match between the underlying data distributions. The best results are in \textbf{bold}.}
\label{tab:zero shot baseline}
\resizebox{\linewidth}{!}{
\begin{tabular}{clccccc|clccccc}
\toprule
Dataset & \makecell{Pred.\\Len.} & Patch. & iTrans. & DLinear & FITS & GPT4TS
&
Dataset & \makecell{Pred.\\Len.} & Patch. & iTrans. & DLinear & FITS & GPT4TS \\
\midrule

\multirow[c]{8}{*}{\makecell{NYTS-h-obs \\ $\rightarrow$ NYTS-h-hid}}
& 24
& \makecell{0.611 \\ \footnotesize{(+0.010)}}
& \makecell{0.606 \\ \footnotesize{(+0.025)}}
& \makecell{0.650 \\ \footnotesize{(+0.012)}}
& \makecell{0.655 \\ \footnotesize{(+0.011)}}
& \makecell{\textbf{0.603} \\ \footnotesize{(+0.021)}}
&
\multirow[c]{8}{*}{\makecell{CCP-h \\ $\rightarrow$ NYTS-h-hid}}
& 24
& \makecell{0.835 \\ \footnotesize{(+0.234)}}
& \makecell{0.866 \\ \footnotesize{(+0.285)}}
& \makecell{\textbf{0.758} \\ \footnotesize{(+0.120)}}
& \makecell{0.763 \\ \footnotesize{(+0.119)}}
& \makecell{0.867 \\ \footnotesize{(+0.285)}} \\

\cmidrule{2-7}
\cmidrule{9-14}

& 168
& \makecell{\textbf{1.123} \\ \footnotesize{(+0.097)}}
& \makecell{1.142 \\ \footnotesize{(+0.101)}}
& \makecell{1.187 \\ \footnotesize{(+0.089)}}
& \makecell{1.228 \\ \footnotesize{(+0.103)}}
& \makecell{1.162 \\ \footnotesize{(+0.097)}}
&
& 168
& \makecell{1.308 \\ \footnotesize{(+0.282)}}
& \makecell{1.670 \\ \footnotesize{(+0.629)}}
& \makecell{\textbf{1.242} \\ \footnotesize{(+0.144)}}
& \makecell{1.281 \\ \footnotesize{(+0.156)}}
& \makecell{1.334 \\ \footnotesize{(+0.269)}} \\

\cmidrule{2-7}
\cmidrule{9-14}

& 336
& \makecell{\textbf{1.310} \\ \footnotesize{(+0.134)}}
& \makecell{1.329 \\ \footnotesize{(+0.134)}}
& \makecell{1.374 \\ \footnotesize{(+0.112)}}
& \makecell{1.440 \\ \footnotesize{(+0.134)}}
& \makecell{1.325 \\ \footnotesize{(+0.131)}}
&
& 336
& \makecell{1.468 \\ \footnotesize{(+0.292)}}
& \makecell{1.928 \\ \footnotesize{(+0.733)}}
& \makecell{\textbf{1.438} \\ \footnotesize{(+0.176)}}
& \makecell{1.722 \\ \footnotesize{(+0.416)}}
& \makecell{1.681 \\ \footnotesize{(+0.487)}} \\

\cmidrule{2-7}
\cmidrule{9-14}

& 720
& \makecell{\textbf{1.424} \\ \footnotesize{(+0.149)}}
& \makecell{1.445 \\ \footnotesize{(+0.150)}}
& \makecell{1.501 \\ \footnotesize{(+0.123)}}
& \makecell{1.602 \\ \footnotesize{(+0.145)}}
& \makecell{1.513 \\ \footnotesize{(+0.184)}}
&
& 720
& \makecell{\textbf{1.549} \\ \footnotesize{(+0.274)}}
& \makecell{2.130 \\ \footnotesize{(+0.835)}}
& \makecell{1.568 \\ \footnotesize{(+0.190)}}
& \makecell{1.890 \\ \footnotesize{(+0.433)}}
& \makecell{5.044 \\ \footnotesize{(+3.715)}} \\

\bottomrule
\end{tabular}
}
\end{table}

\paragraph{Different Generalization Test} Prior benchmarks often employ transfer strategies that force models to transfer across datasets from entirely different domains (e.g., ETT $\rightarrow$ Weather), or across subsets within the same domain but with mismatched sampling frequencies (e.g., ETTh1 $\rightarrow$ ETTm1) \citep{jintime}. However, our distinction between subjects and channels enables a more realistic setup. For a selected dataset, we train models exclusively on the \texttt{-obs} subsets and evaluate them directly on the held-out \texttt{-hid} subsets, simulating a more practical scenario and avoiding cross-dataset transfer.

To compare generalization test strategies, we evaluated both the prior cross-dataset approach and our proposed cross-subject approach on the NYTS dataset. As shown in Table \ref{tab:zero shot baseline}, the cross-dataset setting (CCP $\rightarrow$ NYTS) introduces substantial out-of-distribution shifts, resulting in higher MSE values and model rankings that deviate from in-domain benchmarks. In contrast, our proposed setting (NYTS-obs $\rightarrow$ NYTS-hid) better reflects practical scenarios, such as adding new sensors to an existing network. This intra-domain setting yields lower MSE values and preserves model rankings consistent with the in-domain results in Table \ref{tab:baseline unimodal}. These results suggest that while cross-dataset testing measures robustness under extreme distribution shifts, our subject- and channel-based split provides a more stable and realistic evaluation of intra-domain generalization. Additional results in Appendix \ref{sec: full generalization} show the same pattern.


\subsection{RQ3: Can multimodal forecasting outperform unimodal forecasting?}

\paragraph{Model Comparison} To verify whether multimodal forecasting models can improve performance as they claim, we conducted a comparison and summarized the results in Table \ref{tab:all domain specific multimodal}. GPT4MTS falls behind the best unimodal baseline as it restricts text usage to the historical horizon, treating it merely as a covariate that offers limited foresight. Conversely, FIATS achieves state-of-the-art performance by leveraging text within the prediction horizon, effectively using it as explicit state guidance to direct the forecasting trajectory. This difference demonstrates that the effectiveness of multimodal forecasting hinges on the model itself.

\paragraph{Textual Data Effectiveness Ablation} To explicitly quantify the multimodal gain and isolate the contribution of textual information, we provide an ablation study on effectiveness of textual data. By substituting aligned text with random strings or empty inputs, we observe significant performance degradation, confirming that models actively leverage specific semantic contexts rather than merely reacting to the presence of textual data. Detailed results are provided in Appendix \ref{sec: data ablation}.

\begin{wraptable}{r}{0.5\textwidth}
\centering
\vspace{-0.4cm}
\caption{Domain-specific multimodal forecasting on the full test set. Models are trained on training sets of all datasets. All metrics are MSE. \textcolor{red}{\textbf{Red}}: MSE decrease, \textcolor{blue}{\textbf{Blue}}: MSE increase relative to the best unimodal baseline (Table~\ref{tab:baseline unimodal}).}
\label{tab:all domain specific multimodal}
\resizebox{\linewidth}{!}{
\begin{tabular}{l cccccccc}
\toprule
\multirow{2}{*}{\makecell{Model /\\Dataset}} & \multicolumn{4}{c}{GPT4MTS} & \multicolumn{4}{c}{FIATS} \\
\cmidrule(lr){2-5} \cmidrule(lr){6-9} 
 & 24 & 168 & 336 & 720 & 24 & 168 & 336 & 720 \\
\cmidrule(lr){2-9}
CCP-h & \textcolor{blue}{0.162} & \textcolor{blue}{0.212} & \textcolor{blue}{0.226} & \textcolor{blue}{0.236} & \textcolor{red}{0.109} & \textcolor{red}{0.159} & \textcolor{red}{0.168} & \textcolor{red}{0.183} \\
GREG-h & \textcolor{blue}{0.093} & \textcolor{blue}{0.139} & \textcolor{blue}{0.151} & \textcolor{blue}{0.169} & \textcolor{red}{0.044} & \textcolor{red}{0.065} & \textcolor{red}{0.082} & \textcolor{red}{0.094} \\
JAP-h & \textcolor{blue}{0.358} & \textcolor{blue}{0.543} & \textcolor{blue}{0.605} & \textcolor{blue}{0.656} & \textcolor{red}{0.333} & \textcolor{red}{0.451} & \textcolor{red}{0.500} & \textcolor{red}{0.502} \\
CAISO-h & \textcolor{blue}{0.183} & \textcolor{blue}{0.370} & \textcolor{blue}{0.443} & \textcolor{blue}{0.543} & \textcolor{blue}{0.202} & \textcolor{blue}{0.429} & \textcolor{blue}{0.509} & \textcolor{blue}{0.600} \\
NYTS-h-obs & \textcolor{blue}{0.617} & \textcolor{blue}{1.099} & \textcolor{blue}{1.290} & \textcolor{blue}{1.335} & \textcolor{red}{0.407} & \textcolor{red}{0.644} & \textcolor{red}{0.685} & \textcolor{red}{0.710} \\
\cmidrule(lr){2-9}
& \multicolumn{4}{c}{GPT4MTS} & \multicolumn{4}{c}{FIATS} \\
\cmidrule(lr){2-5} \cmidrule(lr){6-9}
 & 12 & 144 & 288 & 576 & 12 & 144 & 288 & 576 \\
\cmidrule(lr){2-9}
\begin{tabular}[l]{@{}l@{}}BEAR-obs\end{tabular} & \textcolor{blue}{0.177} & \textcolor{blue}{0.591} & \textcolor{blue}{0.874} & \textcolor{blue}{1.096} & \textcolor{red}{0.094} & \textcolor{red}{0.222} & \textcolor{red}{0.298} & \textcolor{red}{0.323} \\
\bottomrule
\end{tabular}
}
\vspace{-1.0cm}
\end{wraptable}

\paragraph{Case Study} To further illustrate how the FIATS model outperforms unimodal models under multimodal inputs, we provide a case study (in Appendix \ref{sec: mm case study}). By leveraging scheduled weather descriptions, FIATS successfully anticipates weather-induced irregular drops in photovoltaic power, whereas PatchTST, relying solely on historical numerical patterns, fails under such non-periodic disturbances. This highlights the value of causally grounded textual inputs for robust forecasting.

\subsection{RQ4: Are Prompting-based LLMs effective for forecasting tasks?}

The rapid improvement in the general capabilities of LLMs has led to their evaluation on specialized time series forecasting tasks. Currently, both general and fine-tuned LLMs typically perform inference via direct prompting; for forecasting tasks, a standard approach involves constructing specific prompts based on forecasting scenario \citep{williams2024context} \citep{wang2024news}. To this end, we designed prompts for unimodal and multimodal contexts. Detailed templates are provided in Appendix \ref{sec:llm prompt}.

\begin{wraptable}{r}{0.5\textwidth}
\centering
\vspace{-0.4cm}
\caption{Performance of LLMs on \texttt{-mini} test set. All metrics are \textbf{MSE}. The best results are highlighted in \textbf{bold} and the second-best are \underline{underlined}. Considering the limitations of LLM context, we choose the \textbf{univariate forecasting datasets} and shorter prediction length (24, 168). "--" indicates that the filled-in prompt exceeds the context limit of the LLM.}
\label{tab: all llms}
\resizebox{\linewidth}{!}{
\begin{tabular}{cll cccccc}
\toprule
\multicolumn{3}{c}{\multirow{2}{*}{Model / Setting}} & \multicolumn{2}{c}{CCP-h-mini} & \multicolumn{2}{c}{GREG-h-mini} & \multicolumn{2}{c}{NYTS-h-mini} \\
\cmidrule(lr){4-5} \cmidrule(lr){6-7} \cmidrule(lr){8-9}
\multicolumn{3}{c}{} & 24 & 168 & 24 & 168 & 24 & 168 \\
\midrule
\multirow{4}{*}{\makecell{Qwen2.5-\\14B-Inst.}} & \multirow{2}{*}{128K} & Uni. & 0.311 & 0.865 & 0.229 & 0.497 & 1.884 & 3.599 \\
 & & Mul. & 0.315 & 0.570 & 0.267 & 0.405 & 2.537 & 3.184 \\
\cmidrule(lr){2-9}
 & \multirow{2}{*}{1M} & Uni. & 0.289 & 0.374 & 0.213 & 0.279 & 1.893 & 2.378 \\
 & & Mul. & 0.303 & 0.555 & 0.274 & 0.421 & 2.562 & 3.061 \\
\midrule
\multirow{2}{*}{\makecell{Qwen3\\-14B}} & \multirow{2}{*}{128K} & Uni. & 0.235 & 0.298 & 0.153 & 0.263 & 2.236 & 2.546 \\
 & & Mul. & 0.222 & 0.336 & 0.210 & 0.282 & 3.169 & 2.911 \\
\midrule
\multirow{2}{*}{\makecell{Deep.-R1}} & \multirow{2}{*}{128K} & Uni. & \underline{0.182} & \underline{0.267} & \textbf{0.113} & \underline{0.213} & 1.071 & \underline{2.154} \\
 & & Mul. & \textbf{0.137} & \textbf{0.216} & \underline{0.130} & \textbf{0.188} & \underline{1.013} & \textbf{1.138} \\
\midrule
\multirow{2}{*}{Chattime} & \multicolumn{1}{c}{\multirow{2}{*}{--}} & Uni. & 0.263 & 0.457 & 0.223 & 0.486 & \textbf{0.998} & 2.312 \\
 & & Mul. & 0.313 & -- & 0.677 & -- & 2.455 & -- \\
\bottomrule
\end{tabular}
}
\vspace{-0.4cm}
\end{wraptable}

\paragraph{Model Comparison} The results in Table \ref{tab: all llms} yield two primary insights. First, among all evaluated LLMs, the best model DeepSeek-R1 consistently achieves superior performance. The more advanced Qwen-3 also outperforms the previous generation Qwen-2.5 in most cases. This indicates that the performance ranking in forecasting tasks closely mirrors that of general-purpose capability, suggesting that robust general reasoning abilities translate effectively to time-series inference. Second, while the fine-tuned Chattime improves upon the Qwen Series and even outperforms DeepSeek-R1 in some cases, it remains unstable, particularly in multimodal scenarios. This leads to the conclusion that existing fine-tuning methods improve specific abilities but offer no stable forecasting advantage. Additional results for multivariate datasets are summarized in Appendix \ref{sec: full llm}.

\paragraph{Modality Comparison} For Qwen and Chattime, the inclusion of multimodal information does not consistently translate into performance improvements; in many cases, the increased complexity and context length appear to burden the models, resulting in degraded predictive ability. Conversely, DeepSeek-R1 effectively leverages the additional semantic information, yielding superior results in multimodal scenarios compared to unimodal scenarios in the majority of test cases. This suggests that the utility of multimodal context is highly dependent on the model's intrinsic reasoning capacity. 

\paragraph{Case Study} To further investigate the source of the forecasting and reasoning capabilities of SOTA LLMs, we conducted a case study on representative examples (in Appendix \ref{sec:llm example}). The study leads to the observation that when dealing with a unimodal scenario, DeepSeek-R1 focuses on statistical extrapolation, extracting historical trends and patterns to predict. When facing multimodal semantic context, DeepSeek-R1 uses its reasoning ability to analyze the relationship between historical weather and the time series to infer future dynamics based on provided future weather data.

\paragraph{Capability Concerns} Although LLMs have shown progress in forecasting tasks with the improvement of their capability, a broader comparison reveals a critical limitation: compared to the evaluation results of dedicated forecasting models on \texttt{-mini} subsets (in Appendix \ref{sec: full mini}), nearly all LLMs we selected lag behind specialized baselines, while only DeepSeek-R1 surpasses some of them. This suggests that previous benchmarks for evaluating the forecasting ability of LLMs may have offered overly simplified tasks with short contexts, or suffered from description leakage and temporal leakage that reduced the difficulty of prediction. In contrast, under our context-rich and leakage-free forecasting setting, LLMs encounter substantially greater challenges.

\paragraph{Reliability Concerns} Beside the forecasting error, we further evaluated the reliability of the LLMs given the observed failure of LLMs. The comprehensive \textbf{pass@3} results are summarized in Appendix \ref{sec: full pass rate}. The results demonstrate that while LLMs maintain high success rates in scenarios characterized by shorter prediction horizons and fewer variables, failure rates escalate substantially when the models are subjected to longer prediction lengths, multivariate forecasting tasks, or inputs enriched with extensive textual context. This highlights the persistent insufficiency of current LLMs in handling long-context, long-term realistic forecasting tasks via direct prompting.

\section{Discussion}

\subsection{Efficiency analysis}

\begin{wrapfigure}{r}{0.5\textwidth}
    \centering
    \vspace{-0.6cm}
    \caption{Comparison of model parameters}
    \includegraphics[width=0.95\linewidth]{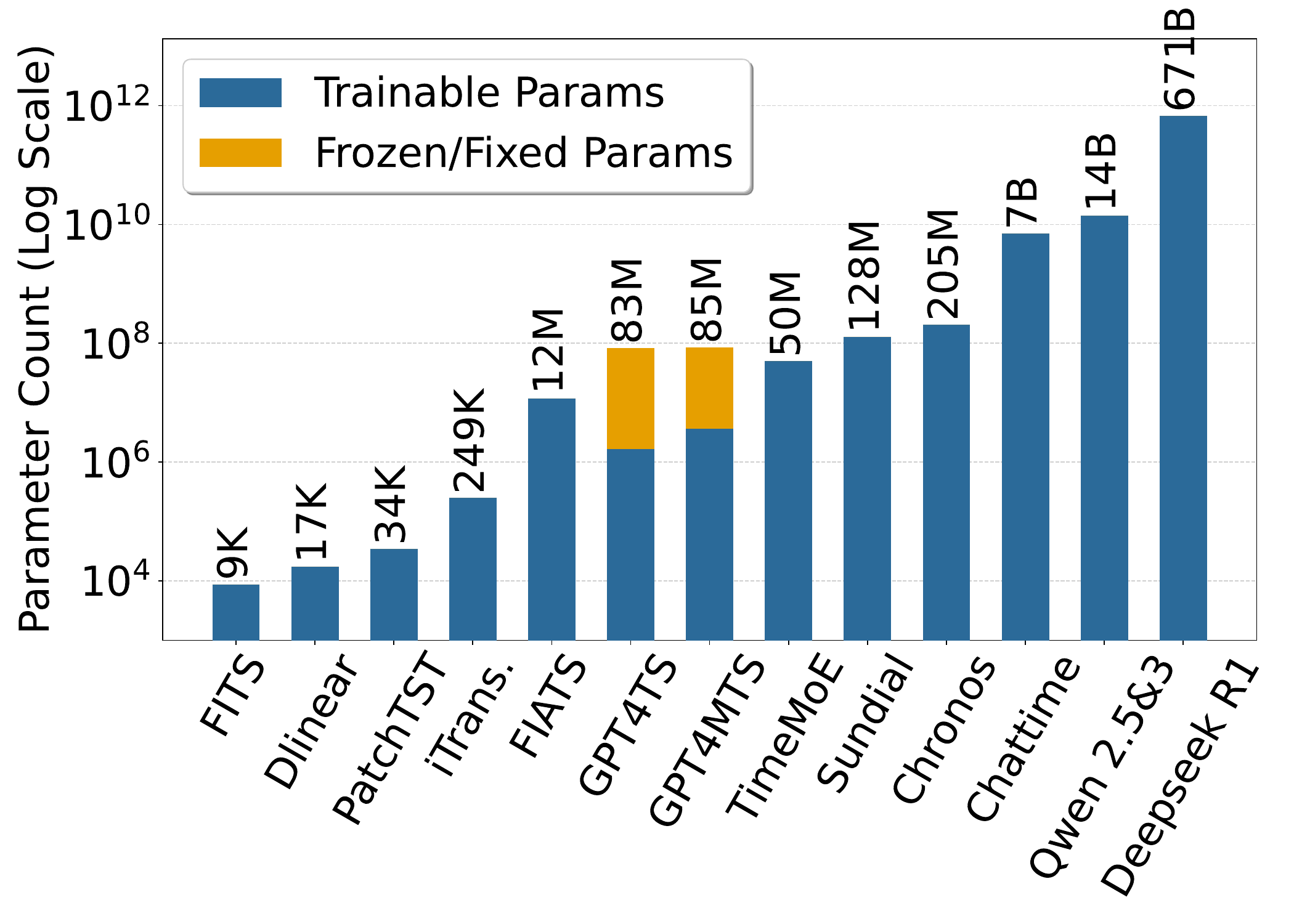}
    \vspace{-0.6cm}
    \label{fig:param comparison}
\end{wrapfigure}

Figure \ref{fig:param comparison} summarizes the parameter statistics. Among domain-specific models, linear models (FITS, DLinear) are the most efficient, followed by transformers, with multimodal FIATS slightly larger than unimodal PatchTST and iTransformer. GPT4TS/GPT4MTS have larger total but limited trainable parameters due to the frozen GPT-2 backbone, with GPT4MTS slightly larger. Parameter scales increase further for pretrained Foundation Models (TimeMoE, Sundial, Chronos) and peak for LLM-based methods, from SFT-based Chattime to general LLMs (Qwen and Deepseek-R1).

\subsection{Limitations}

\paragraph{Trade-off Between High Fidelity Principle and Domain Diversity} The core contribution of our work is the establishment of high fidelity benchmarking principles. To rigorously uphold this standard, we deliberately restrict textual inputs to verifiably scheduled data and focus exclusively on domains where system dynamics are closely tied to such predictable variations. Consequently, we must exclude fields like economics, politics, and healthcare. We acknowledge that this filtering inevitably compromises the generalizability and domain diversity of the benchmark. However, within the strict boundaries of our high fidelity premise, we have curated datasets to encompass as many viable domains as possible, maximizing diversity without ever violating our foundational rules. See Appendix \ref{sec:broader impact} for broader impact discussion.

\section{Conclusion}

In this work, we addressed a critical flaw in the evaluation of time series forecasting models by formalizing the principles of high-fidelity benchmarking and introducing Fidel-TS, a new large-scale benchmark built upon them. Our comprehensive experiments validate Fidel-TS as a rigorous paradigm, revealing a performance landscape different from previous benchmarks. The thorough evaluation of different categories of models reveals new insights.








\newpage

\section*{AI Usage Statement}
We use LLMs solely for language polishing and proofreading during the preparation of this manuscript. The LLMs are not used to generate substantial portions of the manuscript or to contribute to the core ideas, methodology, experiments, analysis, or conclusions. All final content is reviewed and verified by the authors.

\section*{Ethics Statement}
\label{sec:ethic}
All data used in our benchmark was sourced from authentication-protected APIs under publicly accessible license. The datasets contain aggregated sensor readings related to infrastructure and environmental conditions, such as traffic speed, energy generation, and atmospheric physics. No personally identifiable information was collected or used in this study. The data from the UCSD BEAR room dataset, which pertains to rooms within a university building, consists solely of HVAC system measurements and does not include any personal or sensitive information about the occupants.

We acknowledge that misuse of our benchmark may introduce risks, including unfair optimization in sensitive domains, over-specialization to benchmark-specific patterns, and overconfidence in real-world reliability. Benchmark performance should not be taken as evidence of robustness or deployment readiness, especially in safety-critical settings. The benchmark is therefore not intended for automated financial decisions, exploitation of critical infrastructure vulnerabilities, or deployment without rigorous domain-specific validation.

\section*{Reproducibility statement}
Our code and benchmark dataset are publicly accessible through the provided links. We provide the complete data construction pipeline, experimental settings, and evaluation details to facilitate reproduction of our results.

\bibliography{example_paper}
\bibliographystyle{iclr2027_conference}


\newpage
\appendix

\section*{Appendix}

\begingroup 
\color{black}
\hypersetup{linkcolor=black}
\hypersetup{hidelinks}
\startcontents[appendices]
\printcontents[appendices]{l}{1}{\setcounter{tocdepth}{2}}
\endgroup

\section{Additional Benchmark Information}

\subsection{Metadata and Descriptions}
\label{sec:details of datasets}

\paragraph{Canada Calgary Photovoltaics} 
1. General Information: This dataset provides hourly solar power generation data, measured in kilowatt-hours (kWh), from a collection of solar panels located in Calgary, Alberta, Canada.
2. Subjects (9): A Subject in this dataset represents an individual solar panel installed at a specific public facility in Calgary. Each subject is identified by a unique ID (e.g., 314106, 319086). The metadata provides the precise location for each panel, such as "Calgary Fire Hall Headquarters" or "Southland Leisure Centre".
3. Channels (1): Each subject has a single Channel, which is the Solar Power Generation measured in kWh.
4. Textual Data: The primary source of textual information for this dataset is weather data for Calgary, Alberta.

\paragraph{Germany Renewable Energy Grid} 1. General Information: This dataset contains time-series data on the renewable energy supply from four major transmission system operators (TSOs) in Germany. It captures the generation from both solar and wind sources.
2. Subjects (8): Each Subject represents a specific energy source from one of the German power grids. For example, the subject "solar\_50Hertz" represents the solar power generation from the electricity transmission system in the north and east of Germany.
3. Channels (1): Each subject has a single Channel, representing the Energy Generation for that specific source and grid (unit is typically Megawatts, MW).
4. Textual Data: Textual information consists of regional weather data for Germany, aligned to the specific regions served by each TSO, including Bayreuth, Berlin, Dortmund, Stuttgart.

\paragraph{Jena Atmospheric Physics} 1. General Information: This dataset consists of comprehensive atmospheric physical indicators in Jena, Germany. It is a dataset for meteorological time-series forecasting.
2. Subjects (1): There is only one Subject, which represents the entire Jena weather station as a single system.
3. Channels (21): The Channels are the 21 distinct meteorological variables measured by the station. There are some key variables such as: Carbon dioxide concentration (measured in ppm), Actual vapor pressure (measured in mbar).
4. Textual Data: The textual information is sourced from external weather forecast in Jena, Germany.

\paragraph{NYC Traffic Speed} 
1. General Information: The data is collected by traffic sensors deployed across various road segments in New York City, including information about the vehicles on the roads.
2. Subjects (97): A Subject is a unique traffic sensor monitoring a specific road segment in one of the NYC boroughs (e.g., "Queens, with segment of CIP N TNB - Whitestone Expwy S Exit 14 (Linden Pl)"). Each sensor is identified by a unique ID (e.g., 204, 184).
3. Channels (1): Each subject has a single Channel, which is the Average Vehicle Speed in km/h.
4. Textual Data: The textual data is composed of weather information for New York City, covering New York, Queens and Brooklyn.

\paragraph{California ISO} 1. General Information: This dataset provides a holistic view of the California electricity grid, managed by the California Independent System Operator (CAISO).
2. Subjects (1): There is only one Subject, representing the entire CAISO grid system as a single, interconnected entity.
3. Channels (20): The Channels are the 20 distinct metrics that describe the state of the grid. These include demand, power generation from various sources (Hydro, Nuclear, Biomass, etc.), and CO2 emissions.
4. Textual Data: Textual information is derived from state-wide weather data for California, including San Francisco and San Diego.

\paragraph{UCSD BEAR Room} 1. General Information: This dataset contains HVAC (Heating, Ventilation, and Air Conditioning) relevant data collected from 80 rooms within a single building (NAE-01) at the University of California San Diego.
2. Subjects (80): A Subject corresponds to an individual room within the building. Each room is identified by its unique room number (e.g., 104, 208). The metadata describes each room's specific location.
3. Channels (3): Each subject (room) has a consistent set of three Channels: Zone Temperature, Real Power, and Supply Flow.
4. Textual Data: This dataset uniquely features a special type of textual information. It includes internal control event logs (e.g., 'scheduled occupation of a room') from all the rooms.

\begin{table}[ht]
\centering
\caption{Proportion of sensor downtime across different datasets.}
\label{tab:downtime_stats}
\resizebox{0.35\linewidth}{!}{
\begin{tabular}{lc}
\toprule
Dataset & Downtime Proportion (\%) \\
\midrule
CCP     & 0.155\% \\
GREG    & 0.05\% \\
JAP     & 0\%     \\
NYTS    & 0.012\% \\
CAISO   & 0.01\% \\
BEAR    & 0\%     \\
\bottomrule
\end{tabular}
}
\end{table}

\subsection{Statistics of Sensor Downtime}
\label{sec: missing handling}

While our pipeline explicitly models data imperfections such as long missing gaps as ``sensor downtime'' events, the actual proportion of missing values in our raw datasets is extremely small. This minimal missing rate underscores the high quality, reliability, and continuity of the selected authentication-protected API streams. 

Table \ref{tab:downtime_stats} summarizes the percentage of downtime (missing values) for each dataset. As shown, the missing rate is consistently below $0.2\%$ across all benchmarks. This confirms that the overall standard forecasting continuity is preserved, and our model evaluations are not heavily biased by data imputation algorithms.

\subsection{Visualization}
\label{sec:visual}

\begin{figure}
    \centering
    \includegraphics[width=\linewidth]{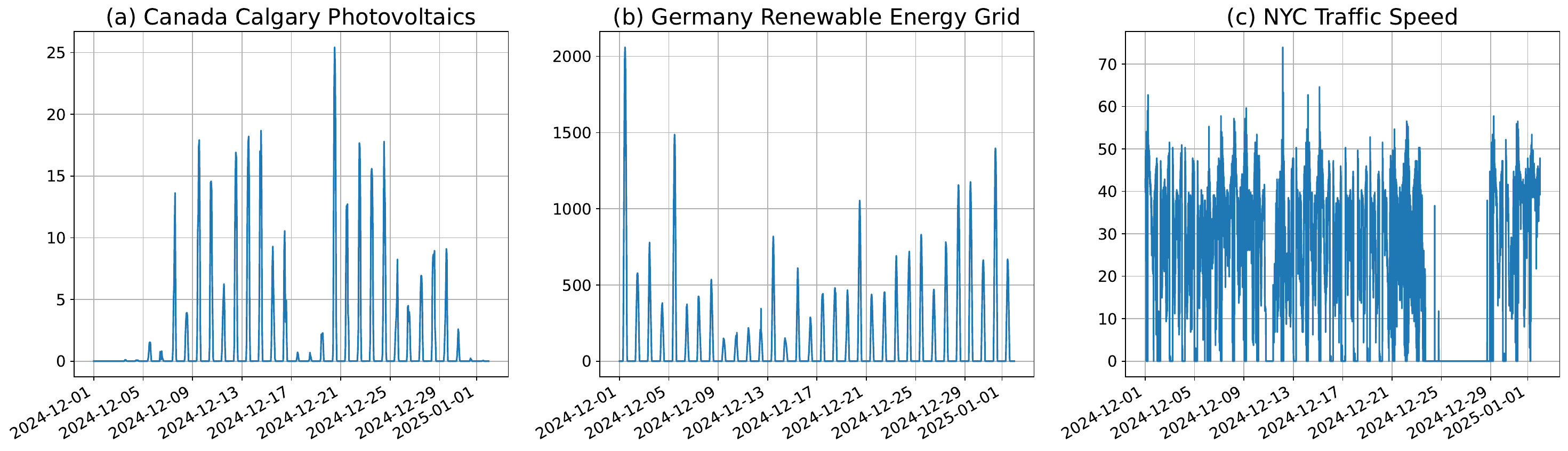}
    \caption{Visualization of the time series data in three datasets with a single channel. We select a representative time periods to show the patterns of the time series in each dataset.}
    \label{fig:visual}
\end{figure}

\paragraph{Time Series Data} For illustrative purposes, we selected datasets with a single channel and plotted one of their subjects over a fixed time range. It can be observed from Figure \ref{fig:visual} that the CCP and GREG datasets exhibit clear periodicity within the selected range. The NYTS dataset is not only noisier and exhibits more frequent, irregular fluctuations, leading to greater prediction difficulty, but also includes periods of sensor downtime, resulting in significant sudden distributional shifts. This suggests that predicting on the NYTS dataset is more challenging and will likely result in higher errors compared to the CCP and GREG datasets.

\begin{figure}
    \centering
    \includegraphics[width=0.7\linewidth]{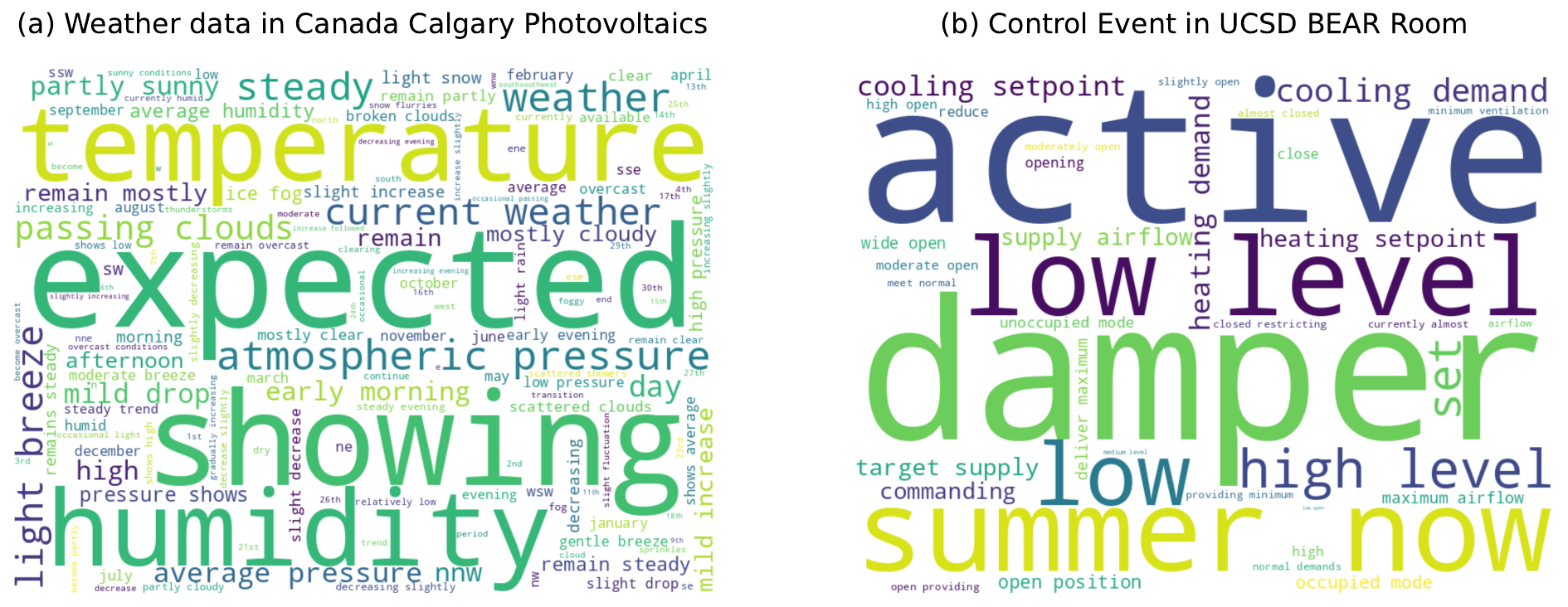}
    \caption{Visualization of the textual data. We select representative weather data and control events.}
    \label{fig:word}
\end{figure}

\paragraph{Textual Data} We also provide word clouds generated from one year of weather data in the CCP dataset and the control events from a single room in the BEAR dataset. As illustrated in Figure \ref{fig:word}, these visualizations clearly highlight the main content of each text source.

\subsection{Importance Sampling Method for LLM evaluation}
\label{sec:importance sampling}

Evaluating LLMs on the full test set presents prohibitive computational costs due to the high frequency and large scale of the data. To address this while ensuring the evaluation remains unbiased, we constructed a representative subset (denoted as \texttt{-mini}) using the importance sampling strategy.

We observed that samples within the test set exhibit varying degrees of multimodal dependency: for some instances, textual information is highly correlated with the time series and significantly aids forecasting, while for others, it offers marginal gains. Simple random sampling risks distorting this distribution. To preserve the original "difficulty" distribution, we employed two lightweight models: a representative unimodal model (PatchTST, \cite{nie2023patch}) and a multimodal model (FIATS, \cite{xu2024intervention}). By evaluating both on the full test set, we categorized samples into two groups based on the performance gap: (1) Text-sensitive samples: Instances where the multimodal model significantly outperformed the unimodal one. (2) Text-insensitive samples: Instances where the improvement was negligible.

Finally, we applied stratified sampling with a fixed ratio to each group. Based on the data volume of each dataset, the final selected ratios are: 1\% per subject for NYTS, 5\% per subject for BEAR, and 10\% per subject for other datasets. This ensures an appropriate data volume for the final -mini test set, while test set faithfully reflects the multimodal sensitivity distribution of the original data, providing a reliable and efficient benchmark for LLM evaluation.

To demonstrate our importance sampling method does not introduce bias into the LLM evaluation, we additionally employ random sampling method to generate subsets and evaluate the LLM. The results are summarized in Appendix \ref{sec:llm random}.

\subsection{Subsetting Method for Generalization Evaluation}
\label{sec:subsetting}
\paragraph{NYTS} The NYTS dataset comprises 97 subjects, representing traffic speed sensors distributed across New York City. Among these, 10 are newly installed sensors with data available only since August 2021. We hold out these 10 subjects to form the \texttt{-hid} subset, while the remaining 87 subjects with complete historical data constitute the \texttt{-obs} subset. This partitioning provides a high-fidelity setting that realistically simulates a common real-world scenario where new sensors are added to an existing network.
\paragraph{BEAR} The BEAR dataset contains complete HVAC data for 80 rooms (subjects) within a single building. We partition the subjects based on their spatial distribution. Specifically, from each floor, we randomly sample a mix of window-side and non-window-side rooms. This process yields an \texttt{-obs} subset of 29 subjects and a \texttt{-hid} subset of 10 subjects.




\section{Framework Implementation Details}
\label{sec:framework details}

Our Unified Cross-modal Evaluation Framework supports both raw text and pre-computed embedding vectors while rigorously enforcing the temporal alignment between textual data and time series. To cater to the varied nature of modern models, our framework not only provides standard PyTorch \citep{paszke2019pytorch} interface, but also integrates HuggingFace Transformers \citep{wolf2020transformers} for foundation models, and leverages PyTorch Lightning \citep{PyTorchLightning_2022} to accelerate the training of complex multimodal models. Recognizing the unique requirements of LLMs, it also supports both local deployment via vLLM \citep{kwon2023efficient} and remote API calls through a simple socket. This thoughtful engineering ensures that our framework serves as the essential bridge between the rich, multimodal data of Fidel-TS and the diverse array of models, enabling the fair and comprehensive experiments that follow.

The key features of our new framework include:

\begin{itemize}
    \item \textbf{Flexible Data Splitting Strategies:} Our framework supports both classical ratio-based partitioning (e.g., 7:1:2 for train/validation/test sets) and exact timestamp-based splitting (e.g., delineating splits at precise dates like 2021-01-01 and 2022-01-01). The timestamp-based approach adheres to chronological order, faithfully simulating realistic forecasting scenarios.
    
    \item \textbf{Hierarchical Subject-Channel Organization:} The framework systematically distinguishes different subjects through unique dataset subset IDs. Within each subject's directory, data is efficiently stored in Parquet files, where the columns are inherently mapped to the distinct channels (i.e., observed variables) associated with that specific subject.
    
    \item \textbf{Customizable Downsampling:} To accommodate varied resolution requirements, the framework incorporates a precise downsampling strategy. Users can specify target frequencies (e.g., 15 minutes or 1 hour), and the system will filter and retain only the data points corresponding to the exact timestamps at the requested intervals.
    
    \item \textbf{Multimodal Integration and Embedding:} The system seamlessly handles multimodal inputs. Textual data can be ingested directly and compressed into vector representations dynamically during testing (utilizing \texttt{jina-embeddings-v3} \citep{sturua2024jina} in our default setup), or loaded as pre-computed embeddings. Furthermore, this embedding-based architecture is inherently extensible, allowing for the future integration of additional modalities (such as images) by compressing them into unified vector spaces.
    
    \item \textbf{Strict Temporal Alignment:} All ingested time-series numerical data and multimodal embeddings are rigorously tagged with exact timestamps. They are consolidated and stored within the Parquet files using perfectly identical timestamp indices, guaranteeing flawless temporal synchronization across all modalities.
    
    \item \textbf{Dynamic Prediction Window Matching:} During the construction of each forecasting sample, the framework dynamically calculates the specific timeframe of the prediction window. It then retrieves the corresponding multimodal embeddings that fall exactly within this window, strictly aligning them by timestamp before feeding the fully synchronized multimodal context into the model.
\end{itemize}


\section{Additional Experiment Information}
\subsection{Model Details}
\label{sec:model}

\begin{table}[ht]
\centering
\caption{Key Information of Models}
\label{tab:all models}
\small
\setlength{\tabcolsep}{4pt}
\begin{tabular}{llllp{3cm}}
\toprule
Model & Training Paradigm & Architecture & Modality & \multicolumn{1}{c}{Textual Input Requirements} \\
\midrule
PatchTST & Domain-Specific & Numerical Transformer & Unimodal & --- \\
iTransformer & Domain-Specific & Numerical Transformer & Unimodal & --- \\
DLinear & Domain-Specific & Linear (MLP) & Unimodal & --- \\
FITS & Domain-Specific & Linear (MLP) & Unimodal & --- \\
GPT4TS & Domain-Specific & LLM-Reprogramming & Unimodal & --- \\
Chronos & Cross-Domain & Numerical Transformer & Unimodal & --- \\
TimeMoE & Cross-Domain & Numerical Transformer & Unimodal & --- \\
Sundial & Cross-Domain & Numerical Transformer & Unimodal & --- \\
GPT4MTS & Domain-Specific & LLM-Reprogramming & Multimodal & History \\
FIATS & Domain-Specific & Numerical Transformer & Multimodal & \makecell[l]{Condition, General Info,\\ Subject/Channel Info} \\
\midrule
\makecell[l]{Qwen2.5-14B-\\Instruct} & Cross-Domain & LLM & \makecell[l]{Unimodal/\\Multimodal} & \makecell[l]{History, Condition,\\ General Info,\\ Subject/Channel Info} \\
\makecell[l]{Qwen2.5-14B-\\Instruct-1M} & Cross-Domain & LLM & \makecell[l]{Unimodal/\\Multimodal} & \makecell[l]{History, Condition,\\ General Info,\\ Subject/Channel Info} \\
\makecell[l]{Qwen3-14B} & Cross-Domain & LLM & \makecell[l]{Unimodal/\\Multimodal} & \makecell[l]{History, Condition,\\ General Info,\\ Subject/Channel Info} \\
\makecell[l]{Deepseek-R1} & Cross-Domain & LLM & \makecell[l]{Unimodal/\\Multimodal} & \makecell[l]{History, Condition,\\ General Info,\\ Subject/Channel Info} \\
\makecell[l]{Chattime} & Cross-Domain & LLM & \makecell[l]{Unimodal/\\Multimodal} & \makecell[l]{Condition, General Info,\\ Subject/Channel Info} \\
\bottomrule
\end{tabular}
\end{table}

Table \ref{tab:all models} contains all the models we select for our experiments. We summarize their Training paradigm, Architecture, Modality and Textual Input Requirements. “General Info" and "Channel/Subject Info” corresponds to the dataset descriptions and the meanings of channels/subjects provided in the metadata. “History” denotes the textual information within the historical window. “Condition” refers to the conditions used for forecasting over the future horizon.

Note that LLM-based domain-specific models (GPT4TS, GPT4MTS) are distinct from LLMs. They only employ a frozen GPT-2 \citep{radford4language} as a backbone, and only the surrounding input and output layers are trainable. Therefore, they must be trained on the dataset and cannot perform zero-shot forecasting.

\subsection{Computational Resources}
\label{sec:computation}
Our experiments were conducted using a computing cluster equipped with 5 NVIDIA L40 GPUs and 2 NVIDIA A100 GPUs. For lightweight MLP-based forecasting models, both training and evaluation can be efficiently performed on CPU resources. Larger Transformer-based architectures require GPU acceleration, which is fully supported by our available infrastructure.

For LLM-based inference, computational requirements are substantially higher. Small- to medium-scale open-source models, including Qwen and Llama series models, can be deployed locally using inference framework. For larger-scale reasoning models such as DeepSeek-R1, local deployment exceeds the available memory capacity, and inference is therefore conducted via API-based access.

\subsection{Metrics}
\label{sec:metrics}
We report the Mean Squared Error (MSE) and Mean Absolute Error (MAE). Let $N$ denote the total number of evaluated sliding windows. 
For the $i$-th test sample, let $\mathbf{y}_{t_i+h} \in \mathbb{R}^{d_y}$ be the ground-truth target vector at prospective step $h \in \{1, \dots, H\}$, and $\hat{\mathbf{y}}_{t_i+h} \in \mathbb{R}^{d_y}$ be the corresponding model prediction generated by $p_\theta$, the formulas are:
\begin{align}
    \mathrm{MSE} &= \frac{1}{N \cdot H} \sum_{i=1}^{N} \sum_{h=1}^{H} \left\| \mathbf{y}_{t_i+h} - \hat{\mathbf{y}}_{t_i+h} \right\|_2^2, \label{eq:mse_def} \\
    \mathrm{MAE} &= \frac{1}{N \cdot H} \sum_{i=1}^{N} \sum_{h=1}^{H} \left| \mathbf{y}_{t_i+h} - \hat{\mathbf{y}}_{t_i+h} \right|_1. \label{eq:mae_def}
\end{align}
Lower values across both metrics indicate superior predictive performance.

\subsection{Settings Details}
\label{sec:settings}
In our experiments, we strictly follow to the principle of Realistic Settings. For all datasets except BEAR, we uniformly adopt the datasets downsampled to hourly (suffixed with \texttt{-h}). For these hourly datasets, we employ a historical horizon of 360 time steps (representing half a month) and prediction lengths of \{24, 168, 336, 720\}, corresponding to \{1 day, 1 week, 2 weeks, and 1 month\}, respectively.

The BEAR dataset, due to its limited time span, is not downsampled to prevent significant data loss. Consequently, we utilize its original 5-minute sampling frequency. The configuration for BEAR includes a historical horizon of 288 time steps (representing 1 day) and prediction lengths of \{12, 144, 288, 576\}, which correspond to \{1 hour, 12 hours, 1 day, and 2 days\}, respectively.

After selecting the appropriate version of each dataset, we partition them into training, validation, and testing sets. We employ the realistic split method based on timestamps. For all datasets except BEAR, we use \{Jan. 1, 2021, Jan. 1, 2022\}, as the split points to separate the three sets. For the BEAR dataset, the corresponding split points were set to \{Sep. 1, 2020, Oct. 1, 2020\}.

Specifically mentioned, for all domain-specific models, both unimodal and multimodal, training must be conducted on each dataset individually. We first train these models and subsequently test them on the full test sets. Following this, we apply our aforementioned importance sampling principle to construct the \texttt{-mini} test sets. For computationally expensive models such as LLMs, evaluation is performed exclusively on these lighter \texttt{-mini} test sets.

\subsection{LLM Prompt Template}
\label{sec:llm prompt}

We provide the prompt templates used for the unimodal and multimodal forecasting tasks with LLMs.

\begin{figure}
    \centering
    \includegraphics[width=1.0\linewidth]{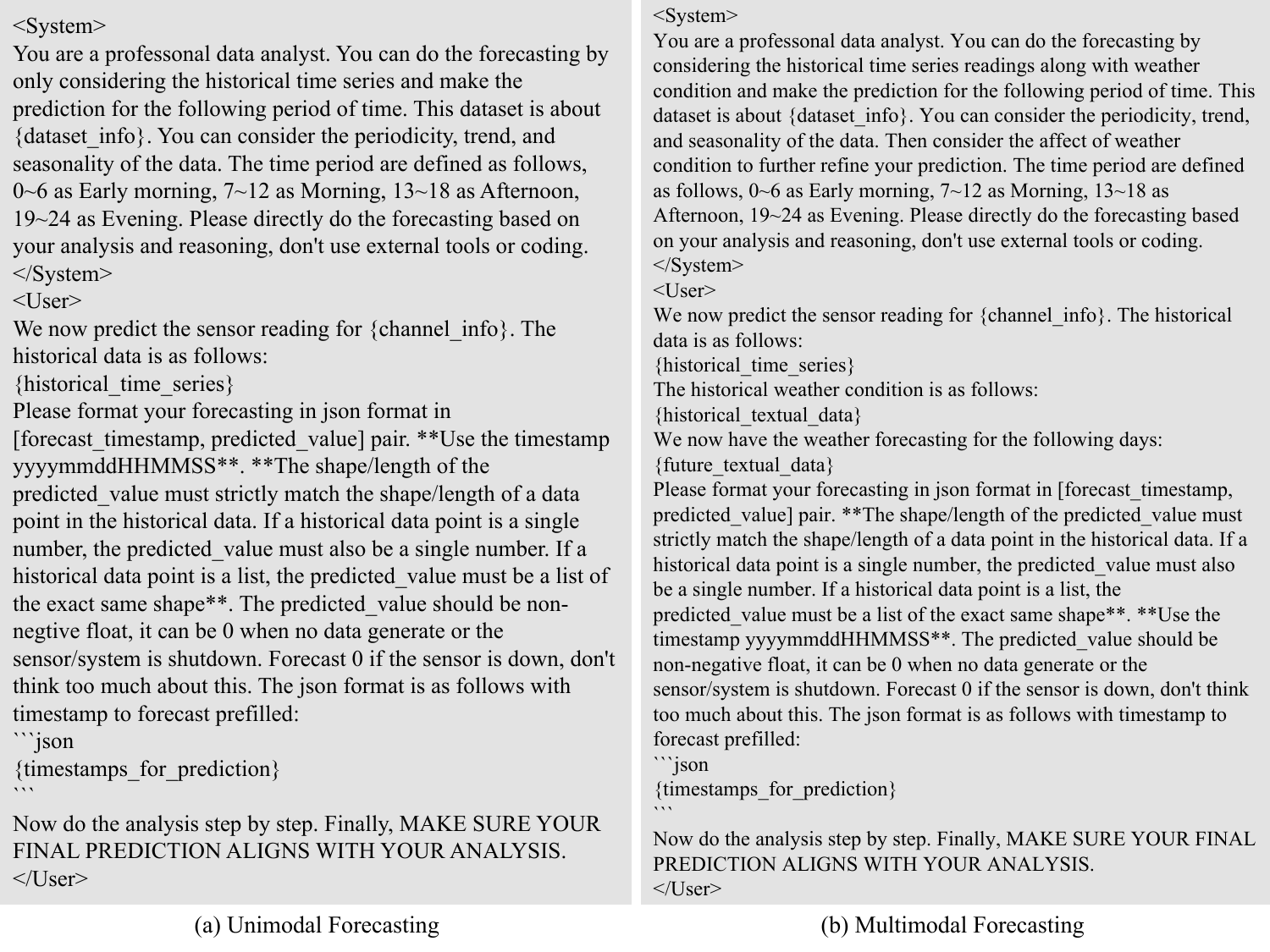}
    \caption{Prompt template for LLM unimodal forecasting and multimodal forecasting}
    \label{fig:prompt}
\end{figure}

The template for unimodal forecasting, illustrated in Figure \ref{fig:prompt} (a), incorporates only basic information about the dataset and channel, along with the formatted historical time series as input. The system prompt directly instructs the model to learn the pattern of time series and generate the forecast.

The multimodal forecasting template shown in Figure \ref{fig:prompt} (b) extends the unimodal setup by additionally providing historical textual information and future conditions in the prediction window. The corresponding system prompt directs the LLM to not only learn the pattern of time series but also focus on the association between the textual data and the time series. For both tasks, we explicitly define the output format to minimize the likelihood of formatting errors.

Notably, we design an additional prompt template for the multimodal forecasting of the BEAR dataset to accommodate its unique control events. The weather-related content is replaced with control-related information, while all other parts remained unchanged. We applied these templates only to native LLMs, including the Qwen series and Deepseek-R1. For the Chattime which fine-tuned Llama-2-7b, we utilized its original template.

\section{Additional Experiment Results}

\subsection{Results of Unimodal Forecasting Models on Datasets without Downsampling}
\label{sec:addtional without downsample}

To maintain the same forecasting setting (a half-month history to predict the next day, the next week, the next 2 weeks and the next month), we adapt the history and prediction lengths to match the native frequency of each dataset. Due to the prohibitive computational cost, we selectively conducted experiments on a subset of models (for transformer-based, MLP-based, and foundation models, we select one representative model from each category) over shorter prediction horizons (day and week), with the results presented in Table \ref{tab:without downsampling}. These results indicate that the models' performance on the raw-resolution data is similar to that on the data downsampled to an hourly frequency.

\begin{table}[h!]
  \centering
  \caption{Performance of selected unimodal forecasting models on full test set of each dataset without downsampling. All metrics are \textbf{MSE}.}
  \label{tab:without downsampling}
  \resizebox{0.6\textwidth}{!}{
    \begin{tabular}{ccccccc}
    \toprule
                          Dataset & \begin{tabular}[c]{@{}c@{}}Samp.\\ Rate\end{tabular} & \begin{tabular}[c]{@{}c@{}}His.\\ Len.\end{tabular} & \begin{tabular}[c]{@{}c@{}}Pred.\\ Len.\end{tabular} & Patch. & DLinear & Sundial \\
    \midrule
    \multirow{2}{*}{GREG}     & \multirow{2}{*}{10min}                               & \multirow{2}{*}{1440}                               & 96                                                   & 0.091 & 0.106   & 0.097   \\
                              &                                                      &                                                     & 672                                                  & 0.131 & 0.154   & 0.166   \\
    \multirow{2}{*}{JAP}      & \multirow{2}{*}{15min}                               & \multirow{2}{*}{2160}                               & 144                                                  & 0.367 & 0.399   & 0.388   \\
                              &                                                      &                                                     & 1008                                                 & 0.557 & 0.565   & 0.610   \\
    \multirow{2}{*}{CAISO}    & \multirow{2}{*}{5min}                                & \multirow{2}{*}{4320}                               & 288                                                  & 0.185 & 0.191   & 0.299   \\
                              &                                                      &                                                     & 2016                                                 & 0.367 & 0.397   & 0.610   \\
    \multirow{2}{*}{NYTS-obs} & \multirow{2}{*}{5min}                                & \multirow{2}{*}{4320}                               & 288                                                  & 0.600 & 0.618   & 0.844   \\
                              &                                                      &                                                     & 2016                                                 & 1.071 & 1.174   & 1.391  \\
    \bottomrule
    \end{tabular}
    }
\end{table}

\subsection{Main Results under the MAE Metric}
\label{sec:all mae}

To ensure a comprehensive evaluation of model performance, we present the forecasting results measured by the Mean Absolute Error (MAE) metric in Table \ref{tab:mae_metric}.

As observed, the relative performance patterns and model rankings under the MAE metric mirror those evaluated under the MSE metric reported in the main text. This demonstrates that the choice between these evaluation metrics is fundamentally robust and does not affect our primary observations and main claims.

\begin{table}[htbp]
    \centering
    \caption{Forecasting results under the MAE metric across various datasets.}
    \label{tab:mae_metric}
    \renewcommand{\arraystretch}{1.1}
    \resizebox{\textwidth}{!}{
    \begin{tabular}{llccccccc}
        \toprule
        Dataset & Pred. Len. & PatchTST & iTrans. & DLinear & FITS & GPT4TS & GPT4MTS & FIATS \\
        \midrule
        \multirow{2}{*}{CCP-h} 
        & 24  & 0.209 & 0.195 & 0.205 & 0.187 & 0.205 & 0.212 & 0.138 \\
        & 168 & 0.240 & 0.220 & 0.228 & 0.216 & 0.233 & 0.243 & 0.197 \\
        \midrule
        \multirow{2}{*}{GREG-h} 
        & 24  & 0.142 & 0.139 & 0.196 & 0.178 & 0.156 & 0.157 & 0.113 \\
        & 168 & 0.169 & 0.167 & 0.281 & 0.214 & 0.205 & 0.190 & 0.145 \\
        \midrule
        \multirow{2}{*}{JAP-h} 
        & 24  & 0.296 & 0.292 & 0.326 & 0.330 & 0.288 & 0.301 & 0.262 \\
        & 168 & 0.408 & 0.412 & 0.400 & 0.406 & 0.406 & 0.410 & 0.318 \\
        \midrule
        \multirow{2}{*}{NYTS-h-obs} 
        & 24  & 0.514 & 0.472 & 0.553 & 0.536 & 0.477 & 0.519 & 0.398 \\
        & 168 & 0.671 & 0.660 & 0.781 & 0.747 & 0.678 & 0.692 & 0.549 \\
        \bottomrule
    \end{tabular}
    }
\end{table}

\subsection{More Generalization Test Results}
\label{sec: full generalization}

\begin{table}[ht]
      \centering
      \caption{Comparison of generalization test on domain-specific unimodal forecasting models. All metrics are \textbf{MSE}. The best results are highlighted in \textbf{bold}.}
      \label{tab:full generalization}
      \resizebox{0.8\linewidth}{!}{
      \begin{tabular}{ll ccccc}
        \toprule
        Dataset & \makecell{Pred.\\Len.} & Patch. & iTrans. & DLinear & FITS & GPT4TS \\
        \midrule
        \multirow[m]{4}{*}{\makecell{BEAR-obs $\rightarrow$ BEAR-hid}} & 12 & 0.100 & \textbf{0.059} & 0.105 & 0.139 & 0.094 \\
        & 144 & 0.424 & \textbf{0.382} & 0.456 & 0.507 & 0.426 \\
        & 288 & 0.588 & \textbf{0.549} & 0.595 & 0.659 & 0.579 \\
        & 576 & 0.762 & \textbf{0.717} & 0.801 & 0.906 & 0.876 \\
        \midrule
        \multirow{2}{*}{CCP-h-obs $\rightarrow$ CCP-h-hid} & 24 & 0.202 & 0.189 & 0.200 & 0.203 & \textbf{0.187} \\
        & 168 & 0.258 & \textbf{0.236} & 0.260 & 0.261 & 0.242 \\
        \midrule
        \multirow{2}{*}{NYTS-h-obs $\rightarrow$ CCP-h-hid} & 24 & 0.229 & \textbf{0.219} & 0.229 & 0.231 & 0.222 \\
        & 168 & 0.276 & \textbf{0.271} & 0.281 & 0.277 & 0.284 \\
        \midrule
        \multirow{2}{*}{GREG-h-obs $\rightarrow$ GREG-h-hid} & 24 & 0.097 & \textbf{0.096} & 0.114 & 0.109 & \textbf{0.096} \\
        & 168 & \textbf{0.133} & 0.135 & 0.163 & 0.135 & 0.136 \\
        \midrule
        \multirow{2}{*}{NYTS-h-obs $\rightarrow$ GREG-h-hid} & 24 & 0.114 & \textbf{0.113} & 0.116 & \textbf{0.113} & \textbf{0.113} \\
        & 168 & 0.148 & 0.150 & 0.174 & \textbf{0.144} & 0.157 \\
        \bottomrule
      \end{tabular}
      }
\end{table}

The generalization test results for the CCP-h and GREG-h datasets are summarized in Table \ref{tab:full generalization}. Consistent with the processing of the NYTS and BEAR datasets, we similarly partition the CCP-h and GREG-h datasets into two disjoint subsets: \texttt{-obs} and \texttt{-hid}. Specifically, for the CCP-h dataset, we allocate 6 subjects to the observed set (CCP-h-obs) for training and the remaining 3 subjects to the hidden set (CCP-h-hid) for generalization evaluation. Similarly, the GREG-h dataset is partitioned by assigning 6 subjects to the observed set (GREG-h-obs) and 2 subjects to the hidden set (GREG-h-hid). 

The experimental results on these two additional datasets remain highly consistent with our findings on the NYTS dataset. The traditional cross-dataset evaluation introduces significant out-of-distribution (OOD) risks, characterized by higher overall forecasting errors and rearranged model rankings compared to their in-domain performance. Conversely, our proposed subject and channel based splitting strategy (\texttt{-obs} $\rightarrow$ \texttt{-hid}) simulates a practical intra-domain transfer scenario. Under this setting, the MSE values are generally lower, and the relative performance rankings of the baseline models are preserved, aligning closely with the in-domain benchmarks.

Moreover, the generalization test results for the BEAR dataset using our proposed strategy are presented in Table \ref{tab:full generalization}. Consistent with the findings on NYTS, the model rankings in this setting remain aligned with those obtained from the in-domain evaluation.

These supplementary results further illustrate the distinct characteristics of the two generalization paradigms. While cross-dataset evaluation highlights the vulnerability of models to massive cross-domain distribution shifts, our splitting method provides a reliable evaluation of how well models generalize to new, unseen entities within a consistent domain environment.

\subsection{Ablation Study on Multimodal Textual Inputs}
\label{sec: data ablation}

To explicitly verify whether multimodal forecasting models truly utilize the semantic information of textual inputs rather than merely treating them as non-specific regularizers, we conduct an ablation study on the FIATS model. We evaluate three distinct input settings across three datasets (CCP-h, GREG-h, and JAP-h):

\begin{itemize}
    \item \textbf{FIATS (Original):} The standard multimodal setting using strictly aligned textual data.
    \item \textbf{FIATS (Random):} The aligned textual data is replaced with text randomly sampled from different time steps, disrupting the causal alignment while maintaining the presence of text tokens.
    \item \textbf{FIATS (All Zero):} The textual input is entirely removed (replaced with empty strings or zero embeddings), forcing the model to rely solely on the numerical time series.
\end{itemize}

As shown in Table \ref{tab:multimodal_ablation}, removing textual information (FIATS all zero) or providing misaligned, random textual information (FIATS random) leads to a significant increase in MSE across all evaluated datasets and prediction lengths. Notably, the random text setting performs similarly to, or occasionally worse than, the all-zero setting. This confirms that the observed multimodal performance gains are derived from the correct semantic correlation between the text and the time series, effectively validating the necessity of high-fidelity, aligned textual data in our benchmark.

\begin{table}[h]
\centering
\caption{Ablation study of multimodal textual inputs on forecasting performance. All metrics are MSE. The best results are highlighted in \textbf{bold}.}
\label{tab:multimodal_ablation}
\resizebox{0.65\linewidth}{!}{
\begin{tabular}{lcccccc}
\toprule
\multirow{2}{*}{Model Setting} & \multicolumn{2}{c}{CCP-h} & \multicolumn{2}{c}{GREG-h} & \multicolumn{2}{c}{JAP-h} \\
\cmidrule(lr){2-3} \cmidrule(lr){4-5} \cmidrule(lr){6-7}
& 24 & 168 & 24 & 168 & 24 & 168 \\
\midrule
FIATS (Original) & \textbf{0.109} & \textbf{0.159} & \textbf{0.044} & \textbf{0.065} & \textbf{0.333} & \textbf{0.451} \\
FIATS (Random)   & 0.150 & 0.262 & 0.112 & 0.129 & 0.431 & 0.558 \\
FIATS (All Zero) & 0.152 & 0.264 & 0.128 & 0.129 & 0.478 & 0.555 \\
\bottomrule
\end{tabular}
}
\end{table}

\subsection{Case Study: Multimodal vs. Unimodal Time Series Forecasting}
\label{sec: mm case study}

\begin{figure}
    \centering
    \includegraphics[width=\linewidth]{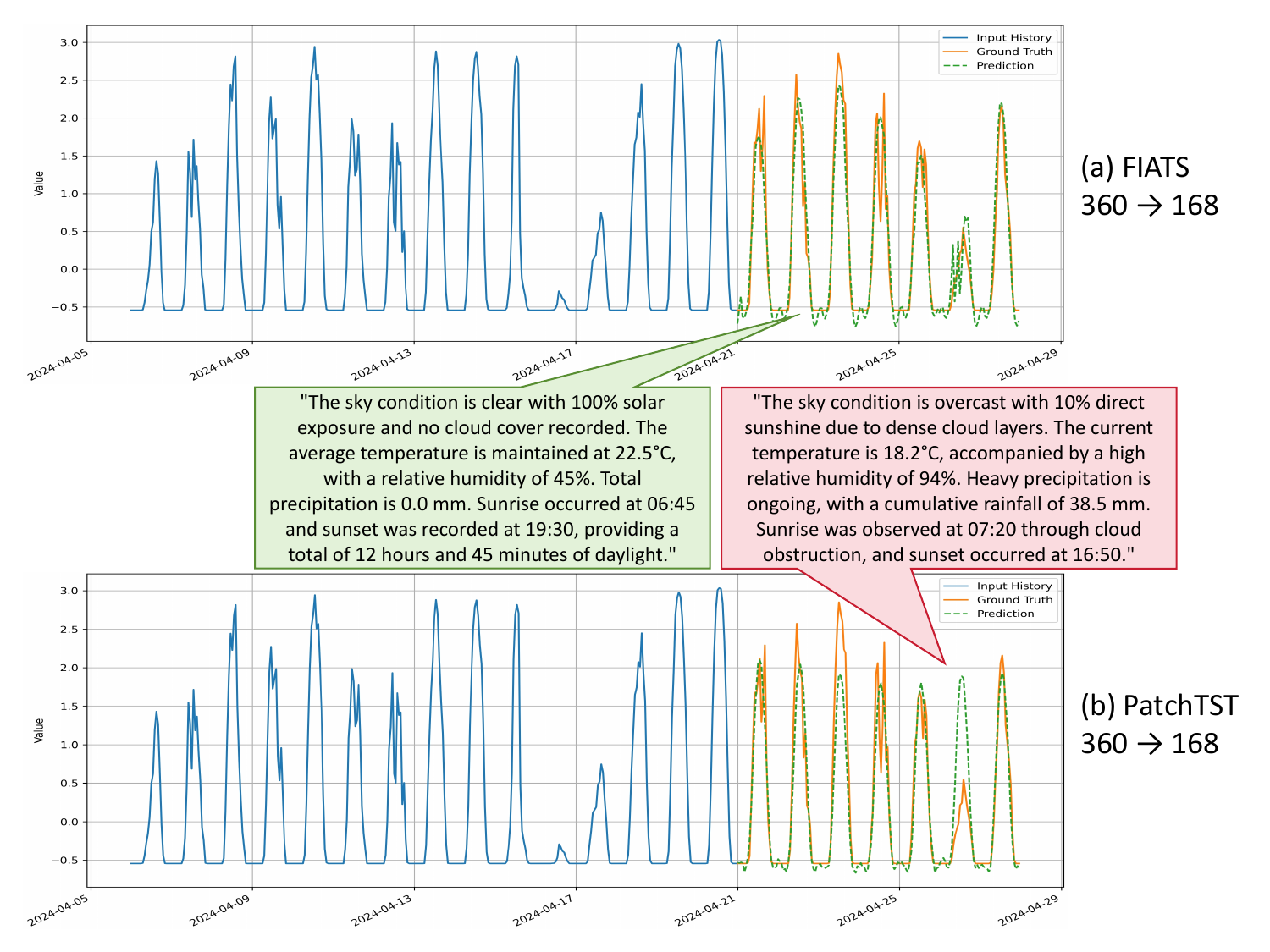}
    \caption{Forecasting comparison between multimodal FIATS and unimodal PatchTST on a sample of CCP-h dataset.}
    \label{fig: mm case study}
\end{figure}

Figure \ref{fig: mm case study} presents a comparative visual analysis of forecasting performance between the multimodal FIATS (a) and the unimodal PatchTST (b). Both panels display the input history (blue solid line), the ground truth (orange solid line), and the respective model predictions (green dashed line). 

As observed in panel (b), the pure time-series model (PatchTST) effectively captures the fundamental diurnal periodicity from the input history. However, it completely misses the significant non-periodic drops in the ground truth that occur around April 26th to 28th, resulting in severe over-prediction.
In contrast, panel (a) demonstrates that FIATS accurately aligns its predictions with these sudden irregular drops. The annotated text boxes illustrate the mechanism behind this success: FIATS leverages context-rich, scheduled textual weather data. When textual inputs indicate ideal conditions (green box: "clear with 100\% solar exposure," "precipitation is 0.0 mm"), the model predicts normal peak values. Crucially, when the scheduled text indicates adverse conditions (red box: "overcast with 10\% direct sunshine," "heavy precipitation," "cumulative rainfall of 38.5 mm"), FIATS correctly utilizes this causally sound information to anticipate a sharp decline in the target variable. This explicitly proves that incorporating scheduled textual modalities enables models to overcome the limitations of relying purely on historical numerical correlations.

\subsection{Evaluation Results of LLMs on Multivariate Datasets}
\label{sec: full llm}

In multivariate forecasting scenarios, the simultaneous processing of numerous variables results in excessive token lengths, frequently causing standard 128k-context LLMs to fail due to context overflow. Consequently, we restrict our evaluation on the multivariate -mini test set to models capable of supporting extended context windows, with the performance results summarized in Table \ref{tab: all llms appendix}. Despite this filtration, we observe two distinct failure modes in these tasks. First, even when operating within theoretical limits, models often struggled to generate the full sequence, resulting in truncated outputs. Second, the overwhelming context length frequently impaired instruction adherence. Models tended to overlook the specific constraint to "predict multivariate variables while maintaining dimensions," resulting in outputs with incorrect feature dimensionality rather than the required original structure.

\begin{table}[h!]
\centering
\caption{Performance of LLMs on \texttt{-mini} test set of multivariate dataset. All metrics are \textbf{MSE}. Considering the limitations of LLM context, we choose the shorter prediction length (24, 168, and for BEAR is (12, 144)). Blank indicates that the filled-in prompt exceeds the input length limit of the LLM.}
\label{tab: all llms appendix}
\resizebox{0.7\linewidth}{!}{
\begin{tabular}{cll cccccc}
\toprule
\multicolumn{3}{c}{\multirow{2}{*}{Model / Setting}} & \multicolumn{2}{c}{\makecell{JAP\\-h-mini}} & \multicolumn{2}{c}{\makecell{CAISO\\-h-mini}} & \multicolumn{2}{c}{\makecell{BEAR\\-mini}} \\
\cmidrule(lr){4-5} \cmidrule(lr){6-7} \cmidrule(lr){8-9} 
\multicolumn{3}{c}{} & 24 & 168 & 24 & 168 & 12 & 144 \\
\midrule
\multirow{2}{*}{\makecell{Qwen2.5-\\14B-Inst.}} & \multirow{2}{*}{1M} & Uni. & 1.825 & -- & 0.501 & -- & 0.522 & 7.425  \\
 & & Mul. & 1.457 & -- & 0.421 & -- & \makecell{--} & -- \\
\midrule
\multirow{2}{*}{Chattime} & \multicolumn{1}{c}{\multirow{2}{*}{--}} & Uni. & 0.483 & 0.580 & 0.216 & 0.453 & 0.145 & 1.077 \\
 & & Mul. & 0.624 & -- & 0.874 & -- & \makecell{--} & -- \\
\bottomrule
\end{tabular}
}
\end{table}

\subsection{Case Study: Example of LLM Reasoning on a Forecasting Task}
\label{sec:llm example}

\begin{figure}
    \centering
    \includegraphics[width=1.0\linewidth]{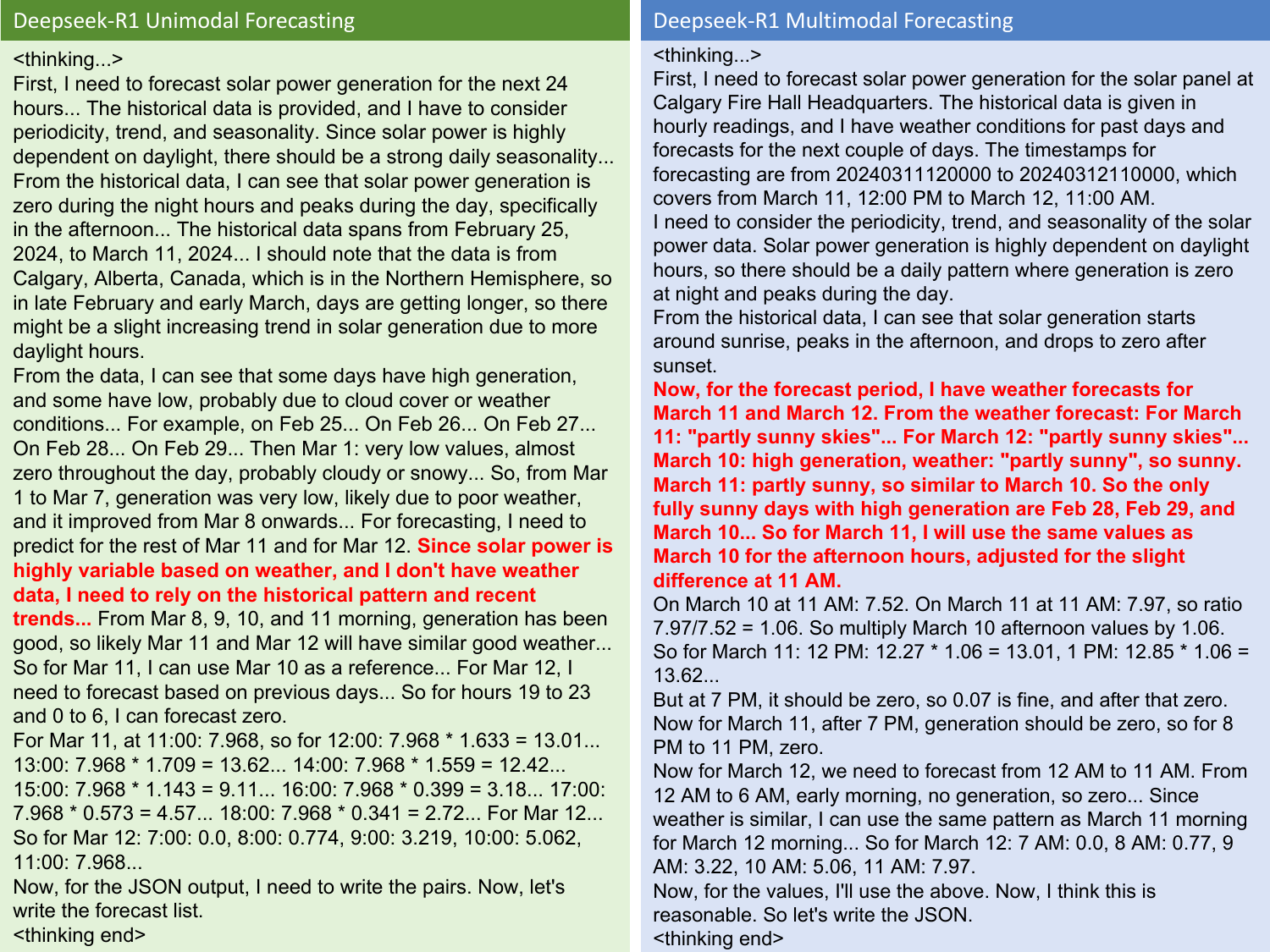}
    \caption{Comparison of the Reasoning Processes of Deepseek-R1 on Unimodal and Multimodal Forecasting Tasks}
    \label{fig:llmoutput}
\end{figure}

To illustrate how LLMs utilize known information within the context in unimodal and multimodal forecasting scenarios, we compared the reasoning processes of Deepseek-R1 under both tasks. The examples we use are derived from the CCP-h-mini test set. The results are shown in Figure \ref{fig:llmoutput}.

In the unimodal forecasting scenario, the model relies exclusively on historical time-series data and implicit domain knowledge. The reasoning process focuses on identifying periodicity (daily solar cycles) and seasonal trends (increasing daylight in the Northern Hemisphere). Crucially, the model explicitly acknowledges the lack of external variables, stating, "I don't have weather data." Consequently, it defaults to a statistical approach, assuming that recent weather will persist, and generates predictions by applying numerical multipliers derived solely from historical correlations.

In contrast, the multimodal forecasting scenario demonstrates a more robust reasoning capability by integrating textual weather data with numerical history. The model actively utilizes the provided weather forecast ("partly sunny skies") to select a specific reference day (March 10) that shares similar conditions. Instead of simple extrapolation, it employs a logic-driven adjustment: it calculates a ratio based on the difference between the current day and the reference day at a known timestamp (11 AM) and applies this scaling factor to forecast the subsequent hours. This highlights the model's ability to ground its predictions in semantic context rather than relying solely on historical data.

\subsection{Full Results of Forecasting Models on \texttt{-mini} Test Sets}
\label{sec: full mini}

To facilitate a direct comparison between LLMs and specialized forecasting models, we evaluate all baseline models on the \texttt{-mini} test sets, as summarized in Table \ref{tab: full mini}. The performance trends and model characteristics observed on this subset align closely with those from the full test sets. This consistency attests to the fairness of our sampling algorithm, confirming that the \texttt{-mini} test sets effectively reduce the number of samples and thereby lower the computational overhead for LLM evaluation, while maintaining representativeness and avoiding the introduction of model-specific biases.

\begin{table}[h!]
\centering
\caption{Performance of all forecasting models on \texttt{-mini} test set of all datasets. All metrics are \textbf{MSE}.}
\label{tab: full mini}

\resizebox{\linewidth}{!}{
\begin{tabular}{ll cccc cccc cc}
\toprule
\multirow{3}{*}{Dataset} & \multirow{3}{*}{\makecell{Pred.\\Len.}} & \multicolumn{4}{c}{\makecell{Domain-Specific\\(Unimodal)}} & \multicolumn{4}{c}{\makecell{Foundation Models\\(Unimodal)}} & \multicolumn{2}{c}{\makecell{Domain-Specific\\(Multimodal)}} \\
\cmidrule(lr){3-6} \cmidrule(lr){7-10} \cmidrule(l){11-12}
& & Patch. & iTrans. & DLinear & FITS & \makecell{GPT\\4TS} & Chronos & \makecell{Time\\-MoE} & Sundial & \makecell{GPT4\\MTS} & FIATS \\
\midrule
\multirow{2}{*}{CCP-h-mini} & 24 & 0.140 & 0.127 & 0.138 & 0.141 & 0.129 & 0.144 & 0.146 & 0.131 & 0.146 & 0.093 \\
& 168 & 0.192 & 0.169 & 0.197 & 0.197 & 0.179 & 0.203 & 0.219 & 0.192 & 0.199 & 0.142 \\
\midrule
\multirow{2}{*}{GREG-h-mini}
& 24 & 0.073 & 0.070 & 0.089 & 0.086 & 0.090 & 0.093 & 0.089 & 0.085 & 0.092 & 0.047 \\
& 168 & 0.135 & 0.135 & 0.150 & 0.137 & 0.143 & 0.170 & 0.162 & 0.154 & 0.156 & 0.074 \\
\midrule
\multirow{2}{*}{JAP-h-mini}
& 24 & 0.286 & 0.285 & 0.334 & 0.334 & 0.278 & 0.277 & 0.278 & 0.286 & 0.275 & 0.234 \\
& 168 & 0.418 & 0.411 & 0.419 & 0.429 & 0.421 & 0.443 & 0.449 & 0.448 & 0.416 & 0.304 \\
\midrule
\multirow{2}{*}{NYTS-h-mini}
& 24 & 0.826 & 0.764 & 0.864 & 0.871 & 0.774 & 0.880 & 0.923 & 0.777 & 0.838 & 0.387 \\
& 168 & 1.538 & 1.599 & 1.560 & 1.615 & 1.749 & 1.876 & 1.675 & 1.797 & 1.643 & 0.700 \\
\midrule
\multirow{2}{*}{CAISO-h-mini}
& 24 & 0.136 & 0.140 & 0.143 & 0.145 & 0.137 & 0.146 & 0.132 & 0.130 & 0.138 & 0.170 \\
& 168 & 0.276 & 0.297 & 0.324 & 0.285 & 0.278 & 0.297 & 0.298 & 0.282 & 0.303 & 0.364 \\
\midrule
\multirow{2}{*}{BEAR-mini}
& 12 & 0.103 & 0.080 & 0.111 & 0.144 & 0.091 & 0.126 & 0.118 & 0.116 & \makecell{0.093} & \makecell{0.066} \\
& 144 & 0.484 & 0.353 & 0.505 & 0.593 & 0.432 & 0.836 & 0.831 & 1.027 & \makecell{0.540} & \makecell{0.211} \\
\bottomrule
\end{tabular}
}
\end{table}

\subsection{Full Results of Pass@3 in LLM evaluation}
\label{sec: full pass rate}

To systematically evaluate the reliability of LLMs in forecasting tasks, we report the pass@3 metric (Table \ref{tab: passrate_comparison}), which reveals critical stability bottlenecks tied to input complexity and context limitations. The results highlight three primary failure mechanisms:

First, the impact of High-Dimensional Data and Context Explosion.
In unimodal settings, LLMs exhibit high stability on datasets with fewer variables (e.g., CCP, GREG, NYTS), maintaining pass rates above 90\%. However, performance collapses on high-dimensional multivariate datasets like JAP and CAISO. Standard 128K-context models yield a 0 pass rate on these tasks. This is attributed to context explosion, where the sheer volume of historical numerical tokens from multiple channels exceeds the model's effective window, preventing valid output generation.

Second, the "Multimodal Tax" on Reliability.
Introducing textual information significantly degrades stability compared to unimodal baselines. Unlike prior works using sparse text, our benchmark involves high-frequency, dense textual data. This additional token load exacerbates the context bottleneck. For instance, on the BEAR dataset, even robust models like DeepSeek-R1 drop from a 99.9\% pass rate (Unimodal) to 0 (Multimodal) as the combined input length of text and historical data overwhelms the context window. Similarly, the domain-finetuned Chattime, while achieving perfect stability (100\%) in unimodal tasks, fails catastrophically in multimodal long-horizon scenarios (dropping to 0.0\%), indicating that current fine-tuning strategies may not sufficiently adapt models to handle such dense multimodal contexts.

Third, the prediction Horizon Sensitivity.
Across all datasets and models, extending the prediction horizon (e.g., from 24 to 168) consistently lowers pass rates. This suggests that LLMs struggle to maintain coherence over long output sequences, often resulting in truncation or format deviations when generating extended forecasts.
In summary, while LLMs show promise in low-resource settings, their reliability is severely compromised in realistic scenarios requiring the simultaneous processing of high-dimensional variables, long-term history, and extensive textual context.

\begin{table}[h!]
  \centering
  \caption{Summarization of all \textbf{pass@3} between Unimodal (Uni.) and Multimodal (Multi.) forecasting for LLMs.}
  \label{tab: passrate_comparison}
  \resizebox{0.9\linewidth}{!}{
    \centering
      \begin{tabular}{ll cccccccccc}
        \toprule
        \multirow{3}{*}{Dataset} & \multirow{3}{*}{\makecell{Pred.\\Len.}} 
        & \multicolumn{4}{c}{Qwen2.5-14B-Instruct} 
        & \multicolumn{2}{c}{Qwen3-14B} 
        & \multicolumn{2}{c}{\multirow{2}{*}{Chattime}} 
        & \multicolumn{2}{c}{DeepSeek-R1} \\
        \cmidrule(lr){3-6} \cmidrule(lr){7-8} \cmidrule(lr){11-12}
        & & \multicolumn{2}{c}{128K} & \multicolumn{2}{c}{1M} & \multicolumn{2}{c}{128K} & \multicolumn{2}{c}{} & \multicolumn{2}{c}{128K} \\
        \cmidrule(lr){3-4} \cmidrule(lr){5-6} \cmidrule(lr){7-8} \cmidrule(lr){9-10} \cmidrule(lr){11-12}
        & & Uni. & Mul. & Uni. & Mul. & Uni. & Mul. & Uni. & Mul. & Uni. & Mul. \\
        \midrule
        \multirow{2}{*}{CCP-h-mini}
          & 24  & 96.3 & 96.8 & 96.9 & 97.1 & 98.8 & 95.6 & 100.0 & 100.0 & 99.9 & 99.8 \\
          & 168 & 75.0 & 71.4 & 73.6 & 81.5 & 90.6 & 77.2 & 100.0 & 0.0   & 97.5 & 94.9 \\
        \midrule
        \multirow{2}{*}{GREG-h-mini}
          & 24  & 96.9 & 98.1 & 97.6 & 99.4 & 96.7 & 94.3 & 100.0 & 99.2  & 99.8 & 99.8 \\
          & 168 & 74.4 & 91.6 & 73.2 & 86.4 & 88.0 & 77.2 & 100.0 & 0.0   & 93.2 & 97.2 \\
        \midrule
        \multirow{2}{*}{JAP-h-mini}
          & 24  & 0.0  & 0.0  & 29.1 & 47.4 & 0.0  & 0.0  & 100.0 & 100.0 & 0.0  & 0.0  \\
          & 168 & 0.0  & 0.0  & 0.0  & 0.0  & 0.0  & 0.0  & 100.0 & 0.0   & 0.0  & 0.0  \\
        \midrule
        \multirow{2}{*}{NYTS-h-mini}
          & 24  & 93.2 & 95.7 & 93.7 & 97.1 & 97.7 & 96.3 & 100.0 & 99.9  & 99.9 & 100.0 \\
          & 168 & 72.9 & 81.8 & 69.4 & 82.4 & 90.0 & 90.0 & 100.0 & 0.0   & 95.3 & 98.2 \\
        \midrule
        \multirow{2}{*}{CAISO-h-mini}
          & 24  & 0.0  & 0.0  & 45.8 & 69.6 & 0.0  & 0.0  & 100.0 & 48.9  & 0.0  & 0.0  \\
          & 168 & 0.0  & 0.0  & 0.0  & 0.0  & 0.0  & 0.0  & 100.0 & 0.0   & 0.0  & 0.0  \\
        \midrule
        \multirow{2}{*}{BEAR-mini}
          & 12  & 96.7 & 0.0  & 95.2 & \makecell{0.0} & 99.8 & 0.0  & 100.0 & \makecell{0.0} & 99.9 & 0.0  \\
          & 144 & 11.6 & 0.0  & 58.1 & 0.0  & 62.8 & 0.0  & 100.0 & 0.0   & 88.4 & 0.0  \\
        \bottomrule
       \end{tabular}
  }
\end{table}

\subsection{LLM Evaluation Results on Randomly Sampled \texttt{-mini} Subsets}
\label{sec:llm random}

We additionally evaluated the models on randomly sampled subsets (i.e., CCP-h-random, GREG-h-random, and NYTS-h-random). As shown in Table \ref{tab:random_sampled}, the performance patterns and relative rankings of the models under the random sampling setting remain consistent with the results obtained via the importance sampling method presented in the main text. This consistency confirms that our subset partitioning methodology provides a robust evaluation and does not introduce any significant bias to the benchmark.

\begin{table}[htbp]
    \centering
    \caption{Evaluation results on randomly sampled subsets.}
    \label{tab:random_sampled}
    \renewcommand{\arraystretch}{1.1}
    \resizebox{0.75\textwidth}{!}{
    \begin{tabular}{lcccccc}
        \toprule
        \multirow{2}{*}{Model} & \multicolumn{2}{c}{CCP-h-random} & \multicolumn{2}{c}{GREG-h-random} & \multicolumn{2}{c}{NYTS-h-random} \\
        \cmidrule(lr){2-3} \cmidrule(lr){4-5} \cmidrule(lr){6-7}
        & 24 & 168 & 24 & 168 & 24 & 168 \\
        \midrule
        PatchTST       & 0.139 & 0.188 & 0.087 & 0.129 & 0.613 & 1.084 \\
        FIATS          & 0.100 & 0.155 & 0.045 & 0.062 & 0.397 & 0.730 \\
        DeepSeek-R1 uni. & 0.185 & 0.255 & 0.151 & 0.211 & 0.783 & 1.543 \\
        DeepSeek-R1 multi. & 0.133 & 0.217 & 0.138 & 0.174 & 0.766 & 0.784 \\
        \bottomrule
    \end{tabular}
    }
\end{table}

\section{Broader Impact}
\label{sec:broader impact}

By revealing the limitations of existing benchmarks and the biased capabilities of some modern models, we encourage the research community to develop more reliable and truly generalizable benchmarking methods. Our extensible framework also provides long-term community value by allowing for the integration of new datasets and models, thus accelerating progress in the field.

The datasets included in our benchmark cover domains critical to public welfare, such as renewable energy management, electricity grid stability, climate change, urban traffic control, and HVAC control. Advances spurred by this benchmark could lead to more efficient energy consumption, reduced carbon emissions, and smarter city infrastructure.

\end{document}